\documentclass{llncs}
\usepackage{graphicx}
\usepackage{amsmath,amssymb} % define this before the line numbering.
\usepackage{color}
\usepackage[width=122mm,left=12mm,paperwidth=146mm,height=193mm,top=12mm,paperheight=217mm]{geometry}

\pdfoutput=1

\newcommand{\Cauchy}{L}
\newcommand{\Momentum}{L}

\newcommand{\IR}{\mathbb{R}}
\newcommand{\IC}{\mathbb{C}}

\begin{document}
% \linenumbers
\pagestyle{headings}
\mainmatter
\def\ECCV18SubNumber{432}  % Insert your submission number here

\title{Newton-Krylov optimization in PDE-constrained diffeomorphic registration parameterized in the space of band-limited
vector fields} % Replace with your title

\titlerunning{Newton-Krylov PDE-constrained LDDMM in the space of band-limited vector fields}

\authorrunning{Monica Hernandez}

\author{Monica Hernandez}
\institute{Computer Sciences Department \\ Aragon Institute on Engieering Research \\ University of Zaragoza \\ mhg@unizar.es}

\maketitle

\begin{abstract}
PDE-constrained Large Deformation Diffeomorphic Metric Mapping is a particularly 
interesting framework of physically meaningful diffeomorphic registration methods. 
Newton-Krylov optimization has shown an excellent numerical accuracy and an 
extraordinarily fast convergence rate in this framework. 
However, the most significant limitation of PDE-constrained LDDMM
is the huge computational complexity, that hinders the extensive use in Computational Anatomy applications.
In this work, we propose two PDE-constrained LDDMM methods parameterized in the space of band-limited 
vector fields and we evaluate their performance with respect to the most related state of 
the art methods.
The parameterization in the space of band-limited vector fields dramatically alleviates the computational 
burden avoiding the computation of the high-frequency components of the velocity fields that would be 
suppressed by the action of the low-pass filters involved in the computation of the gradient and the
Hessian-vector products.
Besides, the proposed methods have shown an improved accuracy with respect to the benchmark methods.
\end{abstract}

\section{Introduction}

% Parrafo 1.
% 1. Optical flow and non-rigid registration ... computer vision and medical image analysis community.
% 2. Physically meaningful transformations problem
% 3. Importance in ... PIV... atmospheric ... and any image where a physical phenomenon underlies the
% motion of the scene
% 4. Diffeomorphic registration and PDE constrained problems
% 5. Incompressibility, models of tumor growth etc

The analysis of scenes with underlying physically meaningful deformations is a very challenging issue 
in numerous scientific domains.
The estimation of the optical flow in pairs of scenes or sequences where a physical phenomenon drives
the motion of the scene have received great attention from the computer vision 
community in the last 
decades~\cite{Ruhnau_07,Ruhnau_07b,Papadakis_07,Papadakis_08,Alvarez_09,Cuzol_09,Heitz_10,Herlin_12,Heas_13,Zhong_17}.
Important areas of application include fluid mechanics, geophysical flow analysis,
atmospheric flow dynamics, meteorology, and oceanography, among others.
The interest of this problem has been extended to the estimation of physically meaningful deformations 
in medical imaging in clinical domains such as 
healthy heart deformation in cardiac imaging series~\cite{Bistoquet_08,Mansi_11}, 
lung motion during respiration~\cite{Vialard_12_MEDIA,Baluwala_13,Papiez_14}, 
or tumor growth~\cite{Hogea_08,Mang_12}, among others.
There are also important clinical domains where the deformation model is not known, 
although there is active research on finding the most plausible transformation among those explained by 
a physical model~\cite{Sotiras_13}.

Particularly interesting is the estimation of physically meaningful diffeomorphic transformations for
Computational Anatomy applications~\cite{Miller_04}.
% such as 
% intra-subject non-rigid registration of longitudinal brain images, 
% or inter-subject registration of brain images from different individuals~\cite{Miller_04}.
Although the differentiability and invertibility of the diffeomorphisms constitute fundamental 
features for Computational Anatomy, the diffeomorphic constraint does not necessarily guarantee 
that a transformation computed with a given method is physically meaningful for the clinical domain of interest. 
In the last decade, PDE-constrained diffeomorphic registration methods have arisen as an appealing paradigm 
for computing diffeomorphisms under plausible physical models~\cite{Hart_09,Vialard_11,Mang_15}.

The first PDE-constrained diffeomorphic registration method was proposed in~\cite{Hart_09}. 
The variational problem in Large Deformation Diffeomorphic Metric Mapping (LDDMM) 
was augmented into a PDE-constrained variational problem with the state equation.
% introducing hard constraints from the PDEs involved in the definition of the physical 
% models.
Gradient-descent was used for the optimization of the problem.
This PDE-constrained LDDMM approach provided the versatility to impose different physical models 
to the computed diffeomorphisms by simply adding the PDEs associated to the problem as hard constraints.

The family of PDE-constrained methods proposed in~\cite{Mang_15,Mang_16,Mang_17,Mang_17b} is specially interesting. 
The authors have used PDE-constrained LDDMM for modeling compressible and incompressible diffeomorphisms,
boundary preserving non-linear Stokes fluid diffeomorphisms,
and mass and intensity preserving diffeomorphisms.
In these methods, the numerical optimization is approached using second-order optimization in the form of 
inexact Newton-Krylov methods.
Despite the excellent numerical accuracy and the extraordinarily fast convergence rate shown by these methods
the computational complexity is huge~\cite{Mang_16_3D}.
This hinders the extensive use of these methods in Computational Anatomy applications.

Fortunately, these PDE-constrained LDDMM methods involve the action of low-pass filters in the optimization 
update equations of the velocities. 
The action of low-pass filters suppresses the high-frequency components of the velocity fields.
Therefore, the numerical implementations using high-resolution velocity fields invest a significant 
part of the computational load in the accurate computation of the high-frequency components of the velocities.
These computations may be omitted since they finally result equal to or nearly equal to zero by the action of 
the low-pass filters.
Zhang et al. have recently argued that the computational complexity of diffeomorphic registration methods 
involving the action of low-pass filters can be alleviated by using a low-dimensional discretization of the 
velocity fields~\cite{Zhang_15,Zhang_17}.
The authors introduced a novel finite-dimensional Lie algebra structure on the space of band-limited vector 
fields. The variational problem in Younes et al.~\cite{Younes_07} was posed in this vector space yielding a 
much more efficient diffeomorphic registration method with accuracy similar to the method formulated in the 
original space.

The purpose of this article is to propose two PDE-constrained LDDMM methods parameterized 
in the space of band-limited vector fields and to evaluate their performance with respect 
to the most related state of the art methods.
The methods depart from the map-based version of PDE-constrained LDDMM~\cite{Hart_09,Mang_15} and
PDE-constrained LDDMM based on the deformation state equation~\cite{Polzin_16}.
For the first time in the literature, we derive the Hessian-vector expressions for 
Newton- and Gauss-Newton- Krylov optimization of~\cite{Hart_09} and~\cite{Polzin_16}.
The gradient and the Hessian-vector expressions derived in this work are computed for the 
band-limited vector field parameterization.
The proposed methods have been implemented in 3D for compressible-Helmholtz and incompressible-$H^1$ 
regularization in the GPU.
The performance has been evaluated using the Non-Rigid Image Evaluation Project database 
(NIREP,~\cite{Song_10}).

In the following, Section~\ref{sec:PDE-LDDMM} reviews the foundations of LDDMM, 
and provides the derivation of the gradient and the Hessian-vector products of the proposed methods 
in the spatial domain.
Section~\ref{sec:BL-PDE-LDDMM} introduces the background on the space of band-limited vector fields and 
provides the expressions of the gradient and Hessian-vector products corresponding to the 
PDE-constrained LDDMM methods parameterized in the space of band-limited vector fields.
% Next, Section~\ref{sec:Implementation} describes the most important implementational details 
% of the proposed methods.
Next, Section~\ref{sec:Results} shows the evaluation results.  
Finally, Section~\ref{sec:Conclusions} gathers the most remarkable conclusions of our work.

\section{PDE-constrained LDDMM}
\label{sec:PDE-LDDMM}

\subsection{Background on LDDMM}

Let $\Omega \subseteq \mathbb{R}^d, d = 2, 3$ be the image domain.
Images are square-integrable functions on $\Omega$.
$I_0$ and $I_1$ denote the source and the target images.
$Diff(\Omega)$ represents the Riemannian manifold of smooth diffeomorphisms on $\Omega$.
$V$ is the tangent space of the Riemannian structure at the identity diffeomorphism ($id$),
where $V$ is a space of smooth vector fields on $\Omega$.
$Diff(\Omega)$ has a Lie group structure, and $V$ is the corresponding Lie algebra.

The Riemannian metric of $Diff(\Omega)$ is defined from the scalar product in $V$ 
\begin{equation}
 \langle v, w \rangle_V = \langle \Cauchy v, w \rangle_{L^2} = \int_\Omega \langle \Cauchy v(x), w(x) \rangle d\Omega, 
\end{equation}
\noindent where $\Cauchy = (Id - \alpha \Delta)^s, \alpha >0, s \in \mathbb{N}$
is the invertible self-adjoint differential operator associated with the differential structure of 
$Diff(\Omega)$.
We denote with $K$ to the inverse of operator $L$.

LDDMM is formulated from the minimization of the energy functional
\begin{equation}
\label{eq:LDDMM}
E(v) =  \int_0^1 \langle \Cauchy v_t, v_t \rangle_{L^2} dt + \frac{1}{\sigma^2} \Vert I_0 \circ (\phi^{v}_{1})^{-1} - I_1\Vert_{L^2}^2. % + \frac{\alpha}{2} \Vert  I_0 - I_1 \circ \phi^{v}_{0,1}\Vert_{L^2}^2.
\end{equation}
\noindent in the space of time-varying smooth flows of velocity fields in $V$, ${v} \in L^2([0,1],V)$.
Given the smooth flow ${v}:[0,1] \rightarrow V$, $v_t:\Omega \rightarrow \IR^{d} \in V$, 
the diffeomorphism $\phi_1^{v}$ is defined as the solution at time $t=1$ to the transport equation 
\begin{equation}
\label{eq:TransportEquation}
\partial_t \phi_t^{v} = v_t \circ \phi_t^{v}  
\end{equation}
\noindent with initial condition $\phi_0^{v} = id$.
The transformation $(\phi^{v}_{1})^{-1}$ computed from the minimum of $E({v})$ is the diffeomorphism that 
solves the LDDMM registration problem between $I_0$ and $I_1$.
LDDMM was originally proposed in~\cite{Beg_05}.

\subsection{PDE-constrained LDDMM based on the state equation}

% \subsubsection{Problem formulation.}

The PDE-constrained LDDMM variational problem is given by the minimization of 
\begin{equation}
\label{eq:LS-LDDMM}
E(v) = \frac{1}{2} \int_0^1 \langle \Cauchy v_t, v_t \rangle_{L^2} dt + \frac{1}{\sigma^2} \Vert m(1) - I_1\Vert_{L^2}^2, 
\end{equation}
\noindent subject to the state equation
\begin{equation}
\label{eq:StateEquation}
\partial_t m(t) + \nabla m(t) \cdot v_t = 0 \textnormal{ in } \Omega \times (0,1],
\end{equation}
\noindent with initial condition $m(0) = I_0$. 
The velocity field flow $v_t$ can be modeled as either a compressible or an incompressible flow of 
a Newtonian fluid through the equation
\begin{equation}
\label{eq:Incompressibility}
\gamma \nabla \cdot v_t = 0 \textnormal{ in } \Omega \times [0, 1],
\end{equation} 
\noindent where $\gamma \in \{0, 1\}$ is the parameter that adjusts the compressibility of the flow.
The compressible PDE-constrained problem was proposed by Hart et al. with gradient-descent 
optimization~\cite{Hart_09}.
Mag et al. introduced the incompressibility constraint and solved the problem using inexact 
Newton-Krylov optimization~\cite{Mang_15}.

% \subsubsection{Gradient and Hessian derivation.}
 
In PDE-constrained LDDMM, the gradient and the Hessian are computed using the method of Lagrange multipliers. 
Thus, we define the Lagrange multipliers $\lambda:\Omega \times [0,1] \rightarrow \IR$ associated with 
the state equation and $p: \Omega \times [0,1] \rightarrow \IR^d$ associated with the incompressibility 
constraint, and we build the augmented Lagrangian
\begin{equation}
\label{eq:LS-LDDMM-Aug}
E_{aug}(v) =  E(v) +  
\int_0^1 \langle \lambda(t), \partial_t m(t) + \nabla m(t) \cdot v_t \rangle_{L^2} dt + 
\int_0^1 \langle p(t), \gamma \nabla \cdot v_t \rangle_{L^2} dt. \nonumber
\end{equation}

\noindent The first-order variation of the augmented Lagrangian yields the expression of the gradient 
\begin{eqnarray}
\partial_t m(t) + \nabla m(t) \cdot v_t = 0 \textnormal{ in } \Omega \times (0,1] \label{eq:FirstOrder1} \\
-\partial_t \lambda(t) - \nabla \cdot ( \lambda(t) \cdot v_t ) = 0 \textnormal{ in } \Omega \times [0,1) \label{eq:FirstOrder2}  \\
\gamma \nabla \cdot v_t = 0 \textnormal{ in } \Omega \times [0,1] \label{eq:FirstOrder3}  \\
(\nabla_v E_{aug}(v))_t = \Momentum v_t + \lambda(t) \cdot \nabla m(t) + \gamma \nabla p(t) \textnormal{ in } \Omega \times [0,1], \label{eq:FirstOrder4}  
\end{eqnarray}

\noindent subject to the initial and final conditions 
$m(0) = I_0$ and $\lambda(1) = -\frac{2}{\sigma^2}(m(1) - I_1) \textnormal{ in } \Omega$. 
In the following, we will recall $m$ as the state variable and $\lambda$ as the adjoint variable.
Equations~\ref{eq:FirstOrder1} and~\ref{eq:FirstOrder2} will be recalled as the state and adjoint equations, respectively.

% The second-order variation of the augmented Lagrangian yields the expression of the Hessian-vector product
% \begin{eqnarray}
% \label{eq:SecondOrderVariation}
% \partial_t \delta m (t) + \nabla \delta m(t) \cdot v_t + \nabla m(t) \cdot \delta v(t) & = 0 \label{eq:SecondOrder1}\\
% -\partial_t \delta \lambda(t) - \nabla \cdot ( \delta \lambda(t) \cdot v_t ) - \nabla \cdot ( \lambda(t) \cdot \delta v(t) ) & = 0 \label{eq:SecondOrder2} \\
% \gamma \nabla \cdot \delta v(t) & = 0
% \end{eqnarray}
% \begin{equation}
% (H_v E_{aug}(\delta v))_t = \Momentum \delta v(t) + \delta \lambda(t) \cdot \nabla m(t) + \lambda(t) \cdot \nabla \delta m(t) + \gamma \nabla \delta p(t)\label{eq:SecondOrder3}  
% \end{equation}
% \noindent subject to 
% $\delta m(0) = 0$ and $\delta \lambda(1) = -\frac{2}{\sigma^2}\delta m(1) \textnormal{ in } \Omega$.
% Equation~\ref{eq:SecondOrder1} corresponds with the incremental state equation.
% Equation~\ref{eq:SecondOrder2} corresponds with the incremental adjoint equation.

% \subsubsection{Alternative gradient and Hessian derivation.}

It should be noticed that the state variable can be computed from 
\begin{equation}
 \label{eq:State}
 m(t) = I_0 \circ \phi(t), 
\end{equation}
where $\phi(t)$ is computed from the equivalent of the transport equation~\cite{Beg_05,Hart_09,Mang_17}
\begin{equation}
\label{eq:DeformationStateEquation}
\partial_t \phi(t) + D \phi(t) \cdot v_t = 0 \textnormal{ in } \Omega \times (0,1],
\end{equation}
\noindent with initial condition $\phi(0) = id \textnormal{ in } \Omega$.
The adjoint variable can be computed from 
\begin{equation}
\label{eq:Adjoint}
\lambda(t) = J(t) \lambda(1) \circ \psi (t), 
\end{equation}
\noindent where $\psi(t) = \phi(t)^{-1}$ and $J(t) = |D \psi(t) |$.

With this approach, the expression of the gradient is given by Equation~\ref{eq:FirstOrder4}, 
where the state and adjoint variables are computed from Equations~\ref{eq:State} and~\ref{eq:Adjoint}
rather than solving the state end adjoint equations.
The transformations $\phi$, $\psi$, and the Jacobian determinant $J$ are computed from the PDEs
\begin{eqnarray}
\partial_t \phi(t) + D \phi(t) \cdot v_t = 0 \textnormal{ in } \Omega \times [0,1) \label{eq:FFirstOrder5} \\
-\partial_t \psi(t) - D \psi(t) \cdot v_t = 0 \textnormal{ in } \Omega \times (0,1] \label{eq:FFirstOrder6} \\
-\partial_t J(t) - v_t \cdot \nabla J(t) = -J(t) \nabla \cdot v_t \textnormal{ in } \Omega \times (0,1] \label{eq:FFirstOrder7} 
\end{eqnarray}
\noindent subject to $\phi(0) = id$, $\psi(1) = id$, and $J(1) = 1$.

The second-order variation of the augmented Lagrangian yields the expression of the Hessian-vector product
\begin{equation}
\label{eq:HessianStateEquation}
(H_v E_{aug}(\delta v))_t = \Momentum \delta v(t) + \delta \lambda(t) \cdot \nabla m(t) + \lambda(t) \cdot \nabla \delta m(t) + \gamma \nabla \delta p(t),   
\end{equation}
\noindent where
\begin{eqnarray}
\delta m(t) = \nabla I_0 \circ \phi(t) \cdot \delta \phi(t) \label{eq:IncrementalState} \\
\delta \lambda(t) = \delta J(t) \lambda(1) \circ \psi(t) + J(t) \nabla \lambda(1) \circ \psi(t) \cdot \delta \psi(t)\label{eq:IncrementalAdjoint}
\end{eqnarray}
\noindent and $\delta \phi$, $\delta \psi$, and $\delta J$ are computed from the first-order variation of Equations~\ref{eq:FFirstOrder5},~\ref{eq:FFirstOrder6}, and~\ref{eq:FFirstOrder7} 
\begin{eqnarray}
\partial_t \delta \phi(t) + D \delta \phi(t) \cdot v_t + D \phi(t) \cdot \delta v(t) = 0 \textnormal{ in } \Omega \times (0,1] \label{eq:SSecondOrder4} \\
-\partial_t \delta \psi(t) - D \delta \psi(t) \cdot v_t - D \delta \psi(t) \cdot \delta v(t) = 0 \textnormal{ in } \Omega \times [0,1) \label{eq:SSecondOrder5} 
\end{eqnarray}
\vspace{-0.5 cm}
\begin{multline}
-\partial_t \delta J(t) - \delta v(t) \cdot \nabla J(t) - v_t \cdot \nabla \delta J(t) = \\
-\delta J(t) \nabla \cdot v_t - J(t) \nabla \cdot \delta v(t) \textnormal{ in } \Omega \times [0,1) \label{eq:SSecondOrder6} 
\end{multline}
\noindent subject to $\delta \phi(0) = 0$, $\delta \psi(1) = 0$, and $\delta J(1) = 0$.

% \color{green}
% The image based variant of Stokes LDDMM was also proposed by Hart et al. with gradient-descent 
% optimization~\cite{Hart_09}.
% In this work, ... we provide the second-order equations from the differenciation of the first-order
% equations.
% ... Alli proponian PDE como una mejora de image based ... Aqui la rescatamos ... 
% Y ademas la extendemos de gradient descent a second-order ... 
% \color{black}

\subsection{PDE-constrained LDDMM based on the deformation state equation}
 
% \subsubsection{Problem formulation.}

This variant of PDE-constrained LDDMM is formulated from the minimization of Equation~\ref{eq:LS-LDDMM}
subject to the deformation state equation
\begin{equation}
\partial_t \phi(t) + D \phi(t) \cdot v_t = 0 \textnormal{ in } \Omega \times (0,1],
\end{equation}
\noindent and the incompressibility constraint
\begin{equation}
\gamma \nabla \cdot v_t = 0 \textnormal{ in } \Omega \times [0, 1].
\end{equation} 
\noindent The compressible PDE-constrained problem was proposed by Polzin et al. with gradient-descent 
optimization~\cite{Polzin_16}.

The Lagrange multipliers are $\rho:\Omega \times [0,1] \rightarrow \IR^d$, associated with 
the deformation state equation, and $p: \Omega \times [0,1] \rightarrow \IR^d$, associated with the incompressibility 
constraint.
The augmented Lagrangian is given by
\begin{equation}
\label{eq:LS-LDDMM-Aug-Polzin}
E_{aug}(v) =  E(v) +  
\int_0^1 \langle \rho(t), \partial_t \phi(t) + D \phi(t) \cdot v_t \rangle_{L^2} dt + 
\int_0^1 \langle p(t), \gamma \nabla \cdot v_t \rangle_{L^2} dt. \nonumber
\end{equation}

The expression of the gradient is given by the first-order variation of the augmented Lagrangian
\begin{eqnarray}
\partial_t \phi(t) + D \phi(t) \cdot v_t = 0 \textnormal{ in } \Omega \times (0,1] \label{eq:PFirstOrder1} \\
-\partial_t \rho(t) - \nabla \cdot (\rho(t) \cdot v_t ) = 0 \textnormal{ in } \Omega \times [0,1) \label{eq:PFirstOrder2} \\
(\nabla_v E_{aug}(v))_t = \Momentum v_t + D \phi(t) \cdot \rho(t) + \gamma \nabla p(t) \textnormal{ in } \Omega \times [0,1] \label{eq:PFirstOrder3}
\end{eqnarray}
\noindent subject to the initial and final conditions 
$\phi(0) = id$, and $\rho(1) = \lambda(1) \cdot \nabla m(1)$. 

From the second-order variation of the augmented Lagrangian, we obtain the expression of the Hessian-vector product 
\begin{eqnarray}
\partial_t \delta \phi(t) + D \delta \phi(t) \cdot v_t + D \phi(t) \cdot \delta v(t) = 0 \label{eq:PSecondOrder1} \\
-\partial_t \delta \rho(t) - \nabla \cdot ( \delta \rho(t) \cdot v_t ) - \nabla \cdot (\rho(t) \cdot \delta v(t)) = 0 \label{eq:PSecondOrder2} \\
(H_v E_{aug}(\delta v))_t = \Momentum {\delta v}_t + D \delta \phi(t) \cdot \rho(t) +  D \phi(t) \cdot \delta \rho(t) + \gamma \nabla \delta p(t) \label{eq:PSecondOrder3}
% + \gamma \nabla \omega(t)\label{eq:SecondOrder3}  
\end{eqnarray}

\noindent subject to $\delta \phi(0) = 0$, $\delta \rho(1) = \delta \lambda(1) \cdot \nabla m(1) + \lambda(1) \cdot \delta \nabla m(1)$.

% \color{green}
% \color{red} ... Esta variante calcula la body force sin necesidad de separar entre grad m y lambda ... 
% Aqui se propone tambien el metodo de segundo orden
% Therefore, PDE constrained LDDMM can be formulated subject to Equation~\ref{eq:DeformationStateEquation}
% instead of Equation~\ref{eq:StateEquation}.
% This deformation based variant of Stokes LDDMM was proposed by Polzin et al. with gradient descent~\cite{Polzin_14}.
% The state variable is given by $\phi$, where
% \color{black}

\subsection{Newton-Krylov optimization}
 
The optimization of PDE-constrained LDDMM problems using Newton-Krylov optimization yields
to the update equation
\begin{equation}
\label{eq:Newton}
 v^{n+1} = v^n - \epsilon \delta v^n,
\end{equation}
\noindent where $\delta v^n$ is computed from preconditioned conjugate gradient (PCG) on the system
\begin{equation}
\label{eq:KKT}
 H_v E_{aug}( \delta v^n) = - \nabla_v E_{aug}(v^n),
\end{equation}
\noindent with preconditioner $K$. 
It should be noticed that the update equation is written on the reduced space $V$, therefore,
the optimization is a reduced space method.

% Algorithms 1 and 2 in the Supplementary Material gather the steps of both PDE-constrained LDDMM methods. 
The resulting algorithms are particularly memory-consuming due to the time sampling used in the solution 
of the PDEs must be dense.
Otherwise, the Courant-Friedrich-Levy (CFL) condition may be violated resulting in numerical 
instabilities and convergence problems.

Since the low-pass filter $K$ appears as the last operation of PCG operations in Equation~\ref{eq:KKT},
the velocity field $v$ does not develop high-frequency components during the update.
This suggests that using high-resolution in the computation of the gradient and the Hessian-vector
products invests useless time and memory in the computation of the high-frequency components, 
motivating our proposed methods.

\section{Band-Limited PDE-constrained LDDMM}
\label{sec:BL-PDE-LDDMM}

\subsection{Background on the space of band-limited vector fields}

Let $\widetilde{\Omega}$ be the discrete Fourier domain truncated with frequency bounds $K_1,$ $\dots,$ $K_d$.
We denote with $\widetilde{V}$ the space of discretized band-limited vector fields on $\Omega$ with these frequency bounds. 
The elements in $\widetilde{V}$ are represented in the Fourier domain as $\tilde{v}: \widetilde{\Omega} \rightarrow \IC^d$,
$\tilde{v}(k_1, \dots, k_d)$, and in the spatial domain as $\iota(\tilde{v}):\Omega \rightarrow \IR^d$,
\begin{equation}
\iota(\tilde{v})(x_1, \dots, x_d) = \sum_{k_1 = 0}^{K_1} \dots \sum_{k_d = 0}^{K_d} \tilde{v}(k_1, \dots, k_d) e^{2 \pi i k_1 x_1} \dots e^{2 \pi i k_d x_d}. 
\end{equation}

\noindent The application $\iota:\widetilde{V} \rightarrow V$ denotes the natural inclusion mapping of $\widetilde{V}$ in $V$.
The aplication $\pi: V \rightarrow \widetilde{V}$ denotes the projection of $V$ onto $\widetilde{V}$.

The space of band-limited vector fields $\widetilde{V}$ has a finite-dimensional Lie algebra structure using the truncated 
convolution in the definition of the Lie bracket~\cite{Zhang_15}.
We denote with $Diff(\widetilde{\Omega})$ to the finite-dimensional Riemannian manifold of diffeomorphisms on $\widetilde{\Omega}$ 
with corresponding Lie algebra $\widetilde{V}$.
The Riemannian metric in $Diff(\widetilde{\Omega})$ is defined from the scalar product
\begin{equation}
 \langle \tilde{v}, \tilde{w} \rangle_{\tilde{V}} = \langle \tilde{\Cauchy} \tilde{v}, \tilde{w} \rangle_{l^2},  
\end{equation}
\noindent where $\tilde{\Cauchy}$ is the projection of operator $\Cauchy$ in the truncated Fourier domain. 
Similarly, we will denote with $\tilde{K}$, $\widetilde{\nabla}$, and $\widetilde{\nabla \cdot}$ to 
the projection of operators $K$, $\nabla$, and $\nabla \cdot$ in the truncated Fourier domain.
In addition, we will denote with $\star$ to the truncated convolution.
 
\subsection{Band-Limited PDE-LDDMM based on the state equation} 

The variational problem is given by the minimization of 
\begin{equation}
\label{eq:BLEnergy}
 E(\tilde{v}) = \frac{1}{2} \int_0^1 \langle \tilde{L}\tilde{v}_t, \tilde{v}_t \rangle_{l^2} dt + \frac{1}{\sigma^2} \Vert m(1) - I_1 \Vert_{L^2}^2
\end{equation}
\noindent subject to 
\begin{eqnarray}
\partial_t m(t) + \nabla m(t) \cdot \iota(\tilde{v}_t) = 0 \textnormal{ in } \Omega \times (0,1] \\
\gamma \nabla \cdot \tilde{v}_t = 0 \textnormal{ in } \Omega \times [0, 1],
\end{eqnarray}
\noindent with initial condition $m(0) = I_0$. 

The expression of the gradient is computed in the space of band-limited vectors yielding
\begin{equation}
\widetilde{(\nabla_{\tilde{v}} E_{aug}(\tilde{v}))_t} = \tilde{\Momentum} \tilde{v}_t + \pi(\lambda(t) \cdot \nabla m(t)) 
+ \gamma \widetilde{\nabla} \tilde{p}(t),
\end{equation}
\noindent where $m$ and $\lambda$ are computed from 
\begin{eqnarray}
m(t) = I_0 \circ \iota(\tilde{\phi}(t)) \\
\lambda(t) = \iota(\tilde{J}(t)) \lambda(1) \circ \iota(\tilde{\psi}(t)), 
\end{eqnarray}
\noindent and $\tilde{\phi}$, $\tilde{\psi}$, and $\tilde{J}$ are computed from the PDEs in the space of band-limited
vector fields
\begin{eqnarray}
\partial_t \tilde{\phi}(t) + \widetilde{D} \tilde{\phi}(t) \star \tilde{v}_t = 0 \textnormal{ in } \Omega \times [0,1) \\
-\partial_t \tilde{\psi}(t) - \widetilde{D} \tilde{\psi}(t) \star \tilde{v}_t = 0 \textnormal{ in } \Omega \times (0,1] \\
-\partial_t \tilde{J}(t) - \tilde{v}_t \star \widetilde{\nabla} \tilde{J}(t) = -\tilde{J}(t) \widetilde{\nabla \cdot} \tilde{v}_t \textnormal{ in } \Omega \times (0,1]. 
\end{eqnarray}

The expression of the Hessian-vector product is computed analogously, 
\begin{eqnarray}
\delta m(t) = \nabla I_0 \circ \iota(\tilde{\phi}(t)) \cdot \iota( \delta \tilde{\phi}(t)) \\
\label{eq:DeltaLambda}
\delta \lambda(t) = \iota(\delta \tilde{J}(t)) \lambda(1) \circ \iota(\tilde{\psi}(t)) + 
\iota(\tilde{J}(t)) \nabla \lambda(1) \circ \iota(\tilde{\psi}(t)) \cdot \iota( \delta \tilde{\psi}(t) ),
\end{eqnarray}
\noindent where $\delta \tilde{\phi}$ and $\delta \tilde{J}$ are computed from the incremental PDEs in the space of band-limited vector fields, 
yielding 
\begin{equation}
\label{eq:BLHessianStateEquation}
\widetilde{ (H_{\tilde{v}} E_{aug}(\delta \tilde{v}))_t} = \Momentum \delta \tilde{v}(t) + 
\pi( \delta \lambda(t) \cdot \nabla m(t) ) + \pi( \lambda(t) \cdot \nabla \delta m(t) ) + \gamma \widetilde{\nabla} \delta \tilde{p}(t).
\end{equation}
 
\subsection{Band-Limited PDE-LDDMM based on the deformation state equation} 

The variational problem is given by the minimization of Equation~\ref{eq:BLEnergy} subject to 
\begin{equation}
\partial_t \tilde{\phi}(t) + \widetilde{D} \tilde{\phi}(t) \star \tilde{v}_t = 0 \textnormal{ in } \Omega \times [0,1)
\end{equation}
\noindent and the incompressibility constraint defined in $\tilde{V}$.
On the one hand, the expression of the gradient is given by
\begin{eqnarray}
\partial_t \tilde{\phi}(t) + \widetilde{D} \tilde{\phi}(t) \star \tilde{v}_t = 0 \textnormal{ in } \Omega \times (0,1]  \\
-\partial_t \tilde{\rho}(t) - \widetilde{\nabla \cdot}  (\tilde{\rho}(t) \star \tilde{v}_t ) = 0 \textnormal{ in } \Omega \times [0,1)  \\
(\nabla_v E_{aug}(v))_t = \tilde{\Momentum} \tilde{v}_t + \widetilde{D} \tilde{\phi}(t) \star \tilde{\rho}(t) + \gamma \widetilde{\nabla} \tilde{p}(t) \textnormal{ in } \Omega \times [0,1] 
\end{eqnarray}

On the other hand, the expression of the Hessian-vector product is given by
\begin{eqnarray}
\partial_t \delta \tilde{\phi}(t) + \widetilde{D} \delta \tilde{\phi}(t) \star \tilde{v}_t + \widetilde{D} \tilde{\phi}(t) \star \delta \tilde{v}(t) = 0  \\
\label{eq:IncrementalDeformationAdjoint}
-\partial_t \delta \tilde{\rho}(t) - \widetilde{\nabla \cdot} ( \delta \tilde{\rho}(t) \star \tilde{v}_t ) - \widetilde{\nabla \cdot} (\tilde{\rho}(t) \star \delta \tilde{v}(t)) = 0 
\end{eqnarray}
\vspace{-0.5 cm}
\begin{equation}
\label{eq:HessianDeformationEquation}
\widetilde{(H_{\tilde{v}}E_{aug}(\delta \tilde{v}))_t} = \tilde{\Momentum} \delta \tilde{v}_t + \widetilde{D} \delta \tilde{\phi}(t) \star \tilde{\rho}(t) +  
\widetilde{D} \tilde{\phi}(t) \star \delta \tilde{\rho}(t) + \gamma \widetilde{\nabla} \delta \tilde{p}(t). 
% + \gamma \nabla \omega(t)\label{eq:SecondOrder3}  
\end{equation}

% ...La diferencia estriba en que mientras que el primero calcula state y adjoint variables en el dominio espacial,
% aqui es integro en el dominio band-limited...

\subsection{Newton- and Gauss-Newton- Krylov optimization in $\widetilde{V}$}

The optimization of PDE-constrained LDDMM problems is performed in $\tilde{V}$ from the update equation
\begin{equation}
\label{eq:BLNewton}
 \tilde{v}^{n+1} = \tilde{v}^n - \epsilon \delta \tilde{v}^n,
\end{equation}
\noindent where $\delta \tilde{v}^n$ is computed using PCG
\begin{equation}
\label{eq:BLKKT}
 \widetilde{(H_{\tilde{v}} E_{aug}( \delta \tilde{v}^n))} = - \widetilde{(\nabla_{\tilde{v}} E_{aug}(\tilde{v}^n))}
\end{equation}
\noindent with preconditioner $\tilde{K}$. 
% The steps of the algorithms are analogous to Algorithms 1 and 2 in the Supplementary Material.

By construction, the Hessian is positive definite in the proximity of a local minimum.
However, it can be indefinite or singular far away from the solution.
In this case, the search directions obtained with PCG are not guaranteed to be descent directions. 
In order to overcome this problem, one can use a Gauss-Newton approximation dropping expressions 
of $H_{\tilde{v}} E( \delta \tilde{v}^n)$ to guarantee that the matrix is definite positive.
For PDE-LDDMM based on the state equation, 
one can drop the term $\iota(\delta \tilde{J}) \lambda(1) \circ \iota(\tilde{\psi})$ from the incremental adjoint variable 
$\delta \lambda$ (Equation~\ref{eq:DeltaLambda}) and the term $\pi( \lambda \cdot \nabla \delta m )$ from the Hessian 
(Equation~\ref{eq:HessianStateEquation}). 
For PDE-LDDMM based on the deformation state equation, the terms $\widetilde{\nabla \cdot} (\tilde{\rho} \star \delta  \tilde{v})$, 
$\pi(\lambda(1) \cdot \delta \nabla m(1))$, and $\widetilde{D} \delta \tilde{\phi} \star \tilde{\rho}$ should be removed from the
incremental deformation adjoint equation (Equation~\ref{eq:IncrementalDeformationAdjoint}), 
the incremental deformation adjoint variable ($\delta \tilde{\rho}(1)$), and the Hessian (Equation~\ref{eq:HessianDeformationEquation}), respectively.

\section{Results}
\label{sec:Results}

In this section, we evaluate the performance of the proposed methods.
As a baseline for the evaluation, we include the results obtained by the methods most related to our work: 
Mang et al. method with the stationary and non-stationary parameterizations~\cite{Mang_15}, and 
Zhang et al. method in the spatial and band-limited domains~\cite{Zhang_15}.

The experiments have been conducted on the Non-rigid Image Registration Evaluation Project database 
(NIREP).
We resampled the images into volumes of size $180 \times 210 \times 180$. % $90 \times 106 \times 90$. %
Registration was carried out from the first subject to every other subject in the database, yielding 15 registrations 
for each method.

The experiments were run on an NVidia GeForce GTX 1080 ti with 11 GBS of video memory and an Intel Core i7 with
64 GBS of DDR3 RAM.
The codes were developed in the GPU with Matlab 2017a using Cuda $8.0$.
For the non-stationary Mang et al. method we developed a hybrid CPU-GPU implementation due to GPU memory constraints.

\subsection{Quantitative results}

\begin{figure*} [!t]
\centering
\scriptsize
\begin{tabular}{c}
\includegraphics[angle = 0, width = 0.75 \textwidth]{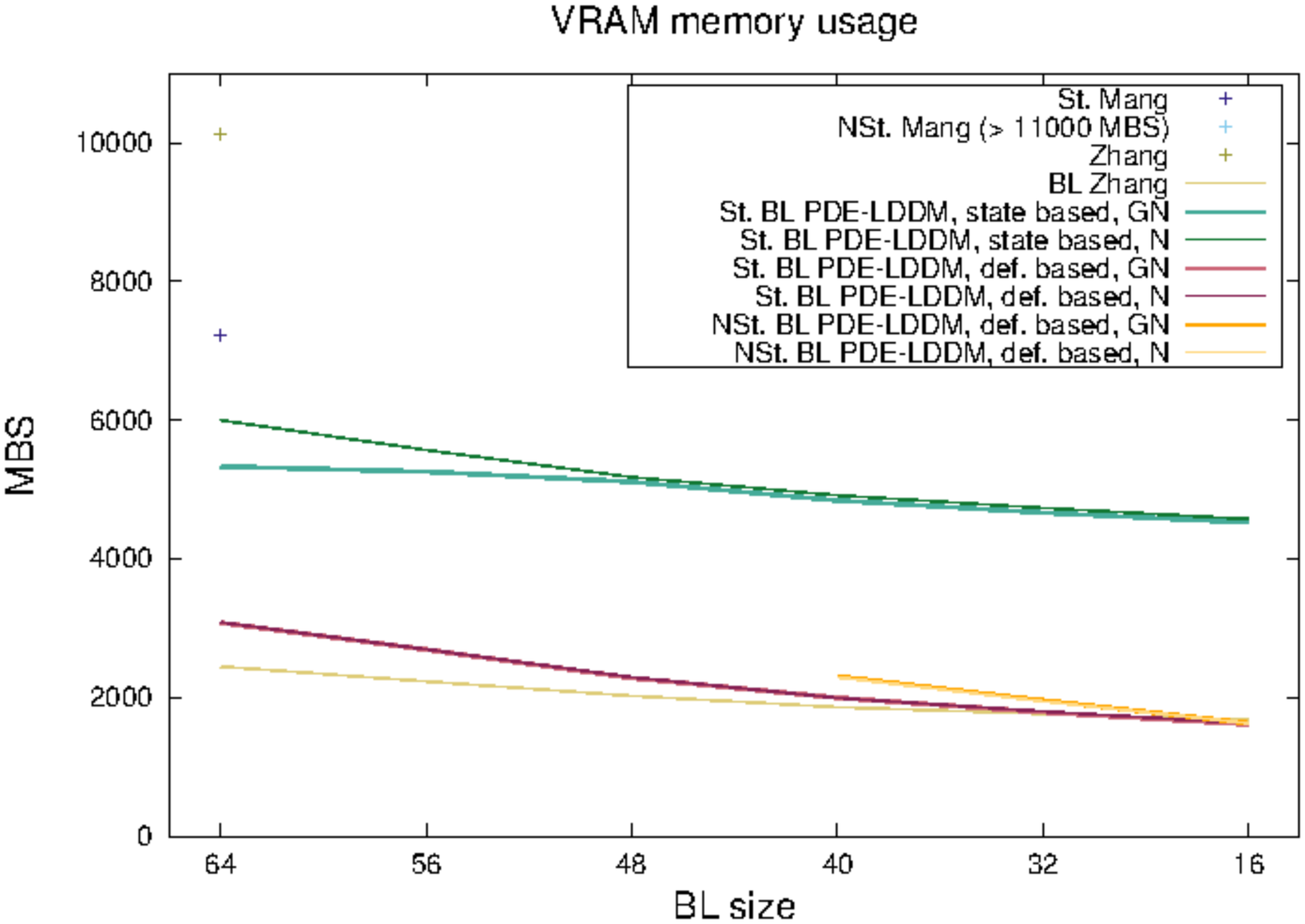} \\
\includegraphics[angle = 0, width = 0.75 \textwidth]{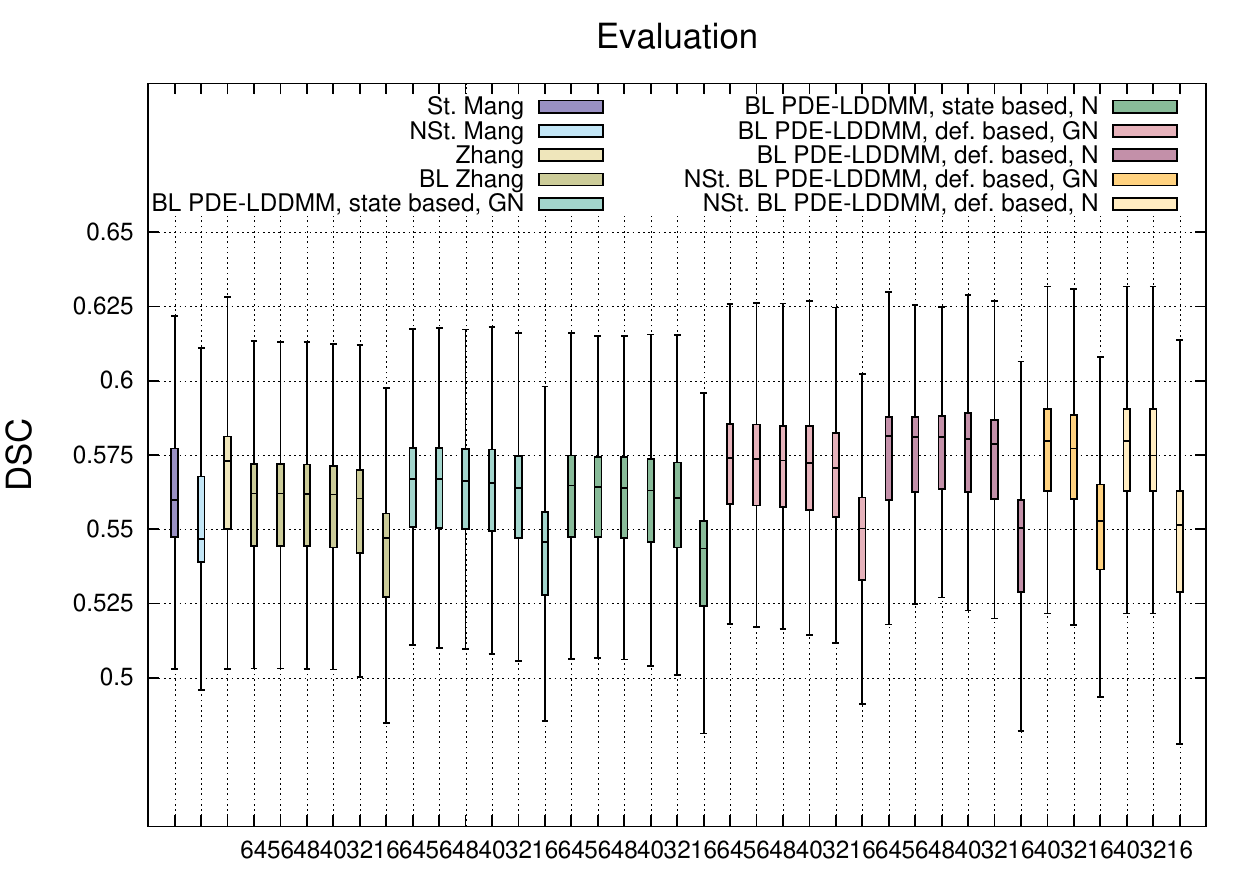}
\end{tabular}
\caption{ \small Up, VRAM peak memory (MBs) reached through the experiments by the registration methods.
Down, volume overlap obtained by the registration methods measured in terms of the DSC between
the warped and the corresponding manual target segmentation. Box and whisker plots show the distribution 
of the DSC averaged over the 32 NIREP manual segmentations. The tick labels show the size of the band-limited
space when applicable.} 
\label{fig:DSC}
\end{figure*}

Table~\ref{table:QuantitativeResultsV} shows 
the quantitative results of most interest for the evaluation of the proposed methods.
The experiments were performed with compressible $V$-regularization. 
In particular, the tables show the relative image similarity error after registration, $MSE_{rel}$,
the relative gradient magnitude, 
$\Vert g \Vert_{\infty,rel} = \frac{\Vert \nabla_{\tilde{v}} E({\tilde{v}}^n) \Vert_\infty}{\Vert \nabla_{\tilde{v}} E({\tilde{v}}^0) \Vert_\infty}$, 
the extrema of the Jacobian determinant, the number of inner PCG iterations, and the total computation time.
Figure~\ref{fig:DSC} left shows the VRAM peak memory reached through the computations.

As most remarkable, our proposed methods outperformed the benchmark in terms of the $MSE_{rel}$ metric with the exception 
of BL sizes equal to $16$.
PDE-LDDMM based on the deformation state equation converged to $MSE_{rel}$ values smaller than the $MSE_{rel}$ values obtained by  
PDE-LDDMM based on the state equation.
Newton and Gauss-Newton optimization converged to similar $MSE_{rel}$ values for all 
BL size configurations.
The use of the non-stationary parameterization improved the obtained $MSE_{rel}$ values.
The relative gradient was reduced to average values ranging from 0.1 to 0.01, 
which means that the optimization was stopped in acceptable energy values for our application.
All Jacobians remained above zero.
Concerning the computational time, our proposed methods outperformed the benchmark methods in the spatial domain,
Gauss-Newton resulted much more efficient than Newton optimization, and the use of the non-stationary parameterization
considerably increased the computational complexity.
Concerning the VRAM memory usage, our PDE-LDDMM based on the deformation state equation was the most efficient 
second-order method, with a memory usage close to the memory needed by BL Zhang gradient-descent. 

\begin{figure*} [!t]
\centering
\scriptsize

\begin{tabular}{ccc}
\hspace{-0.5 cm}
\includegraphics[angle = 0, width = 0.25 \textwidth]{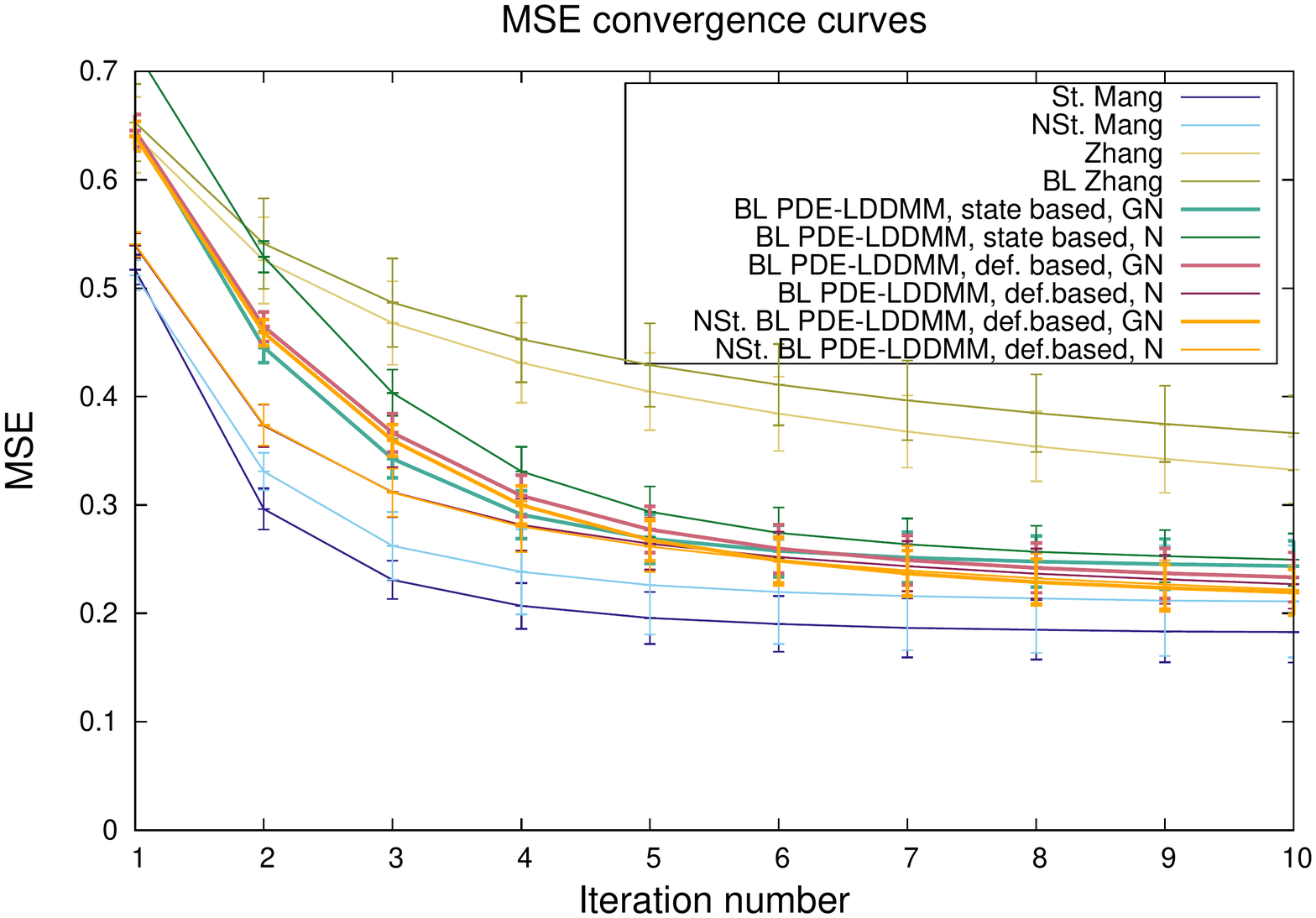} 
&
\includegraphics[angle = 0, width = 0.25 \textwidth]{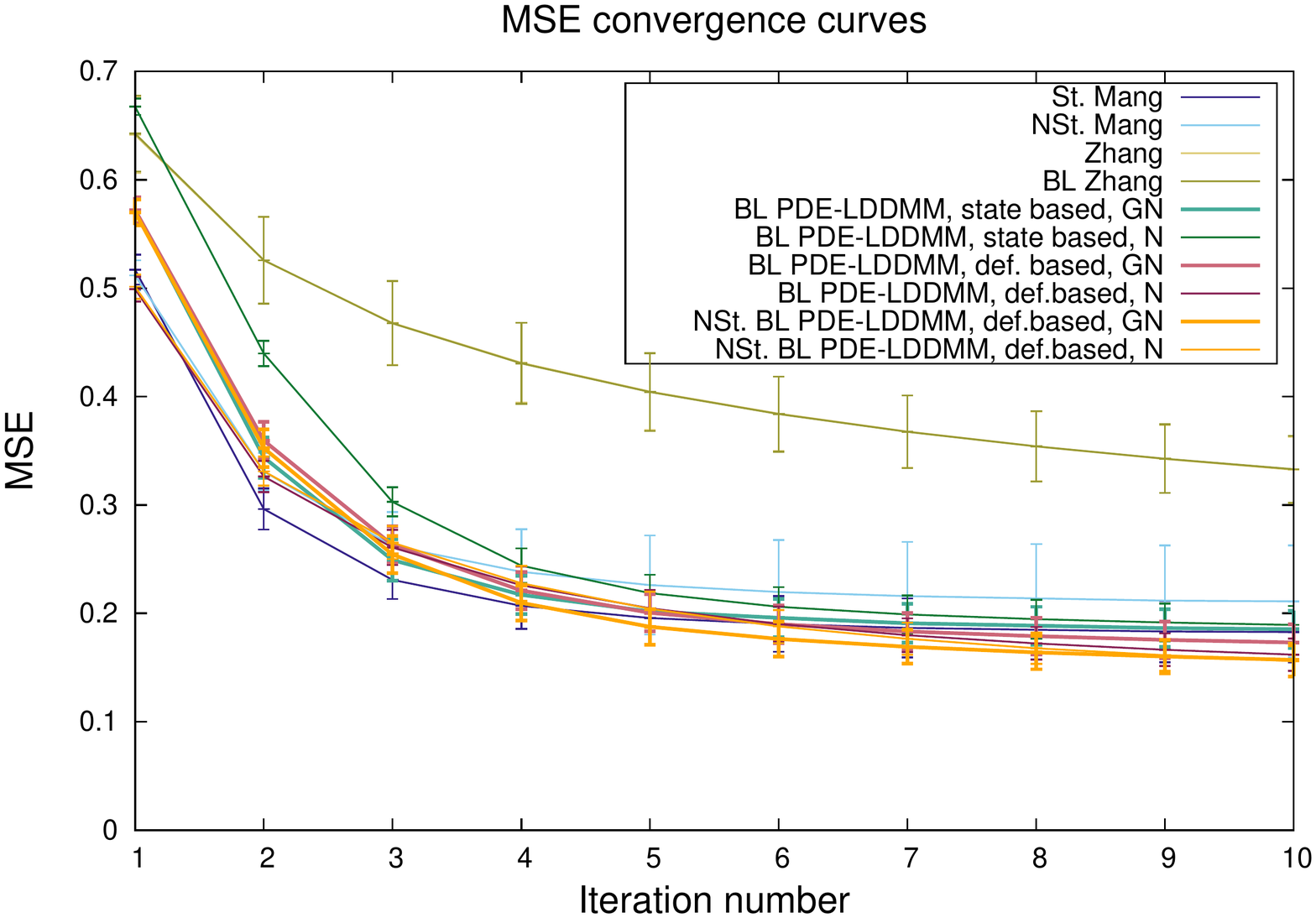} 
&
\includegraphics[angle = 0, width = 0.25 \textwidth]{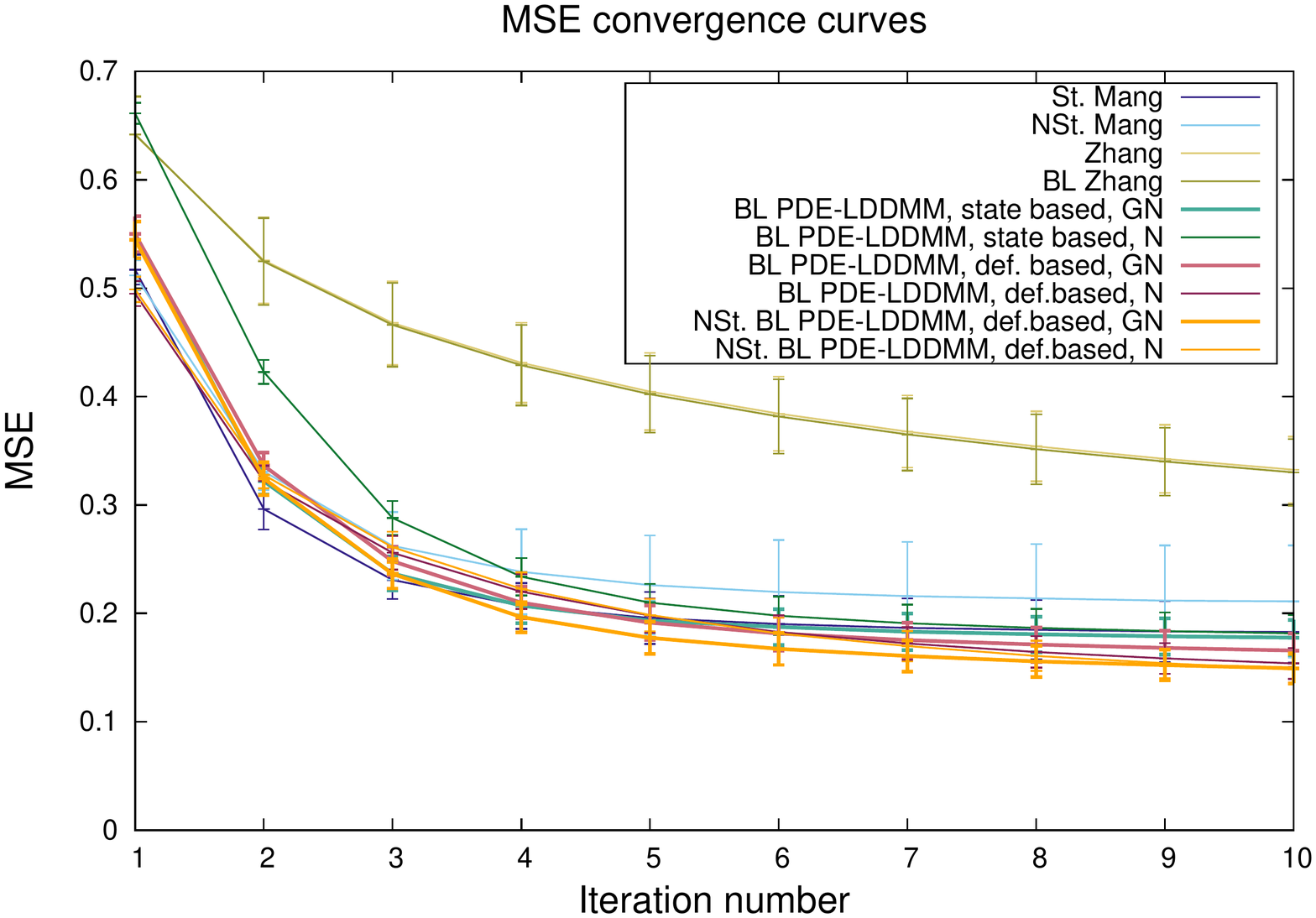} 
\\
\hspace{-0.5 cm}
\includegraphics[angle = 0, width = 0.25 \textwidth]{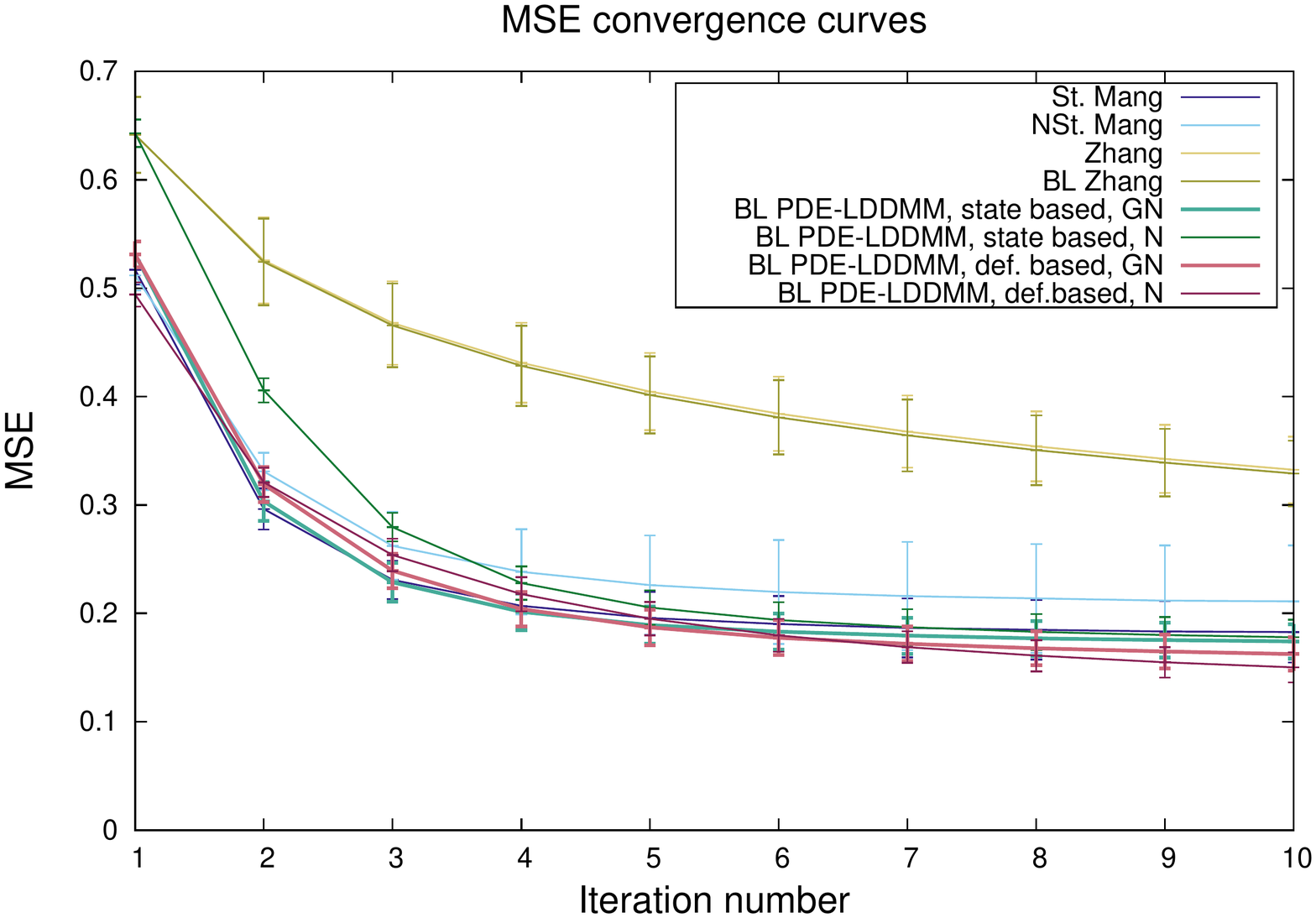} 
&
\includegraphics[angle = 0, width = 0.25 \textwidth]{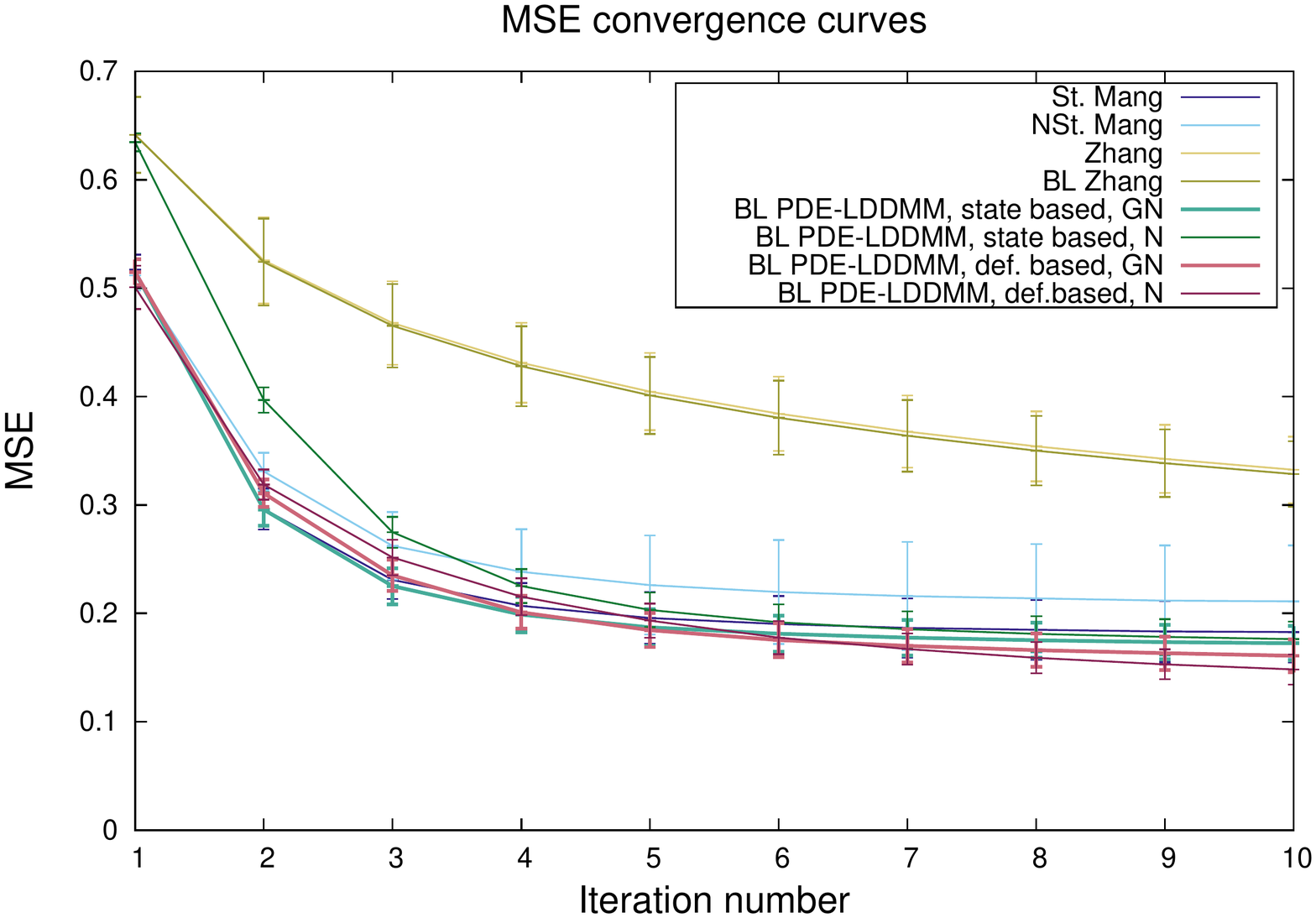} 
&
\includegraphics[angle = 0, width = 0.25 \textwidth]{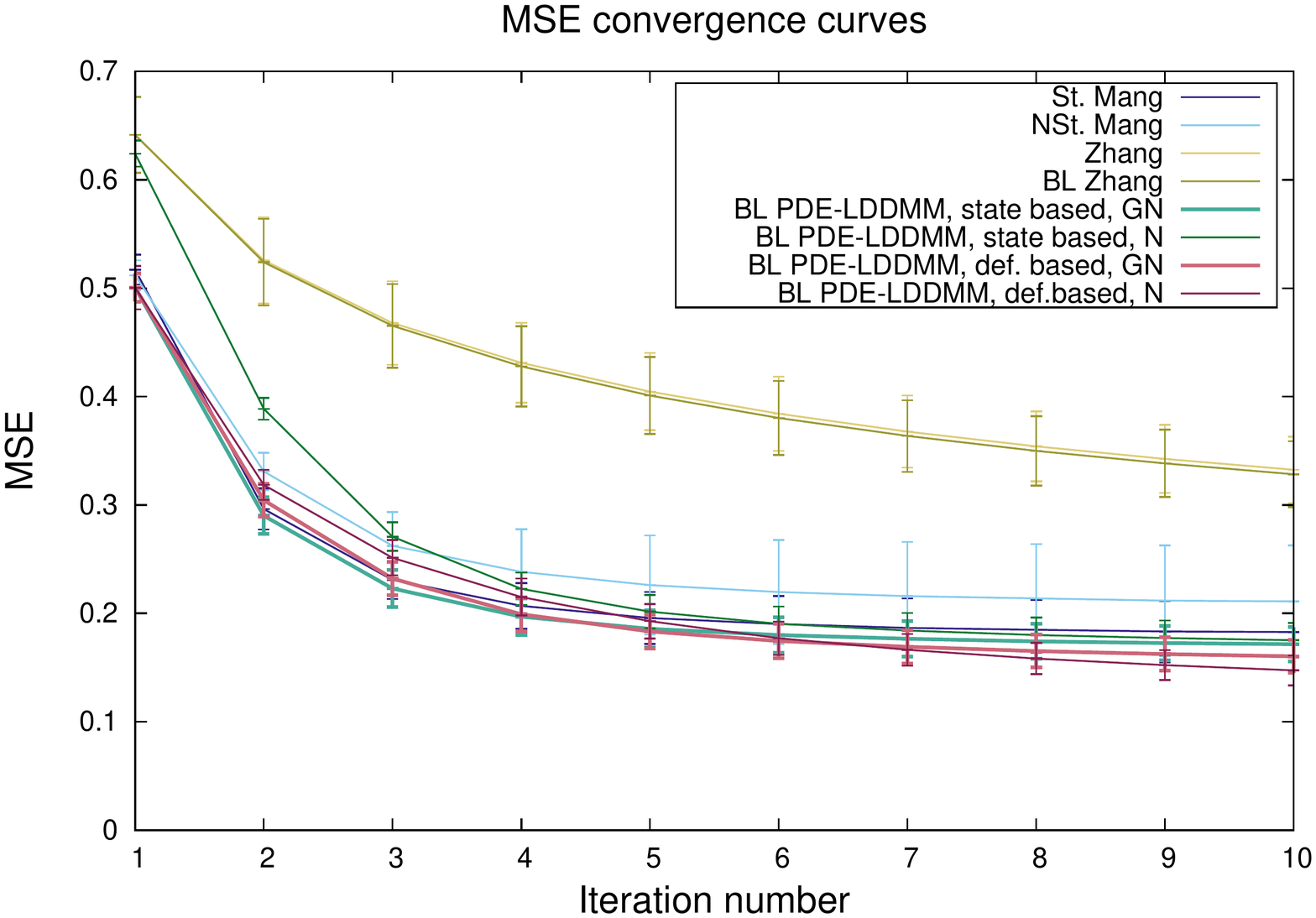} 
\end{tabular}
\caption{ \small $MSE_{rel}$ convergence curves of the proposed methods for each BL domain size.
For comparison purposes, the convergence curves of the benchmark methods in $\Omega$ are shown 
in all the plots.} 
\label{fig:ConvergenceCurves}
\end{figure*}

Figure~\ref{fig:ConvergenceCurves} shows the mean and standard deviation of the $MSE_{rel}$ convergence curves 
obtained during the optimization.
The proposed methods with Gauss-Newton optimization showed a convergence pattern similar to stationary 
Mang. et al. method, with the exception of BL sizes equal to $16$ where the spatial methods showed a faster convergence.
Second-order optimization methods outperformed gradient-descent as theoretically expected.
Newton optimization in PDE-LDDMM based on the deformation state equation outperformed Gauss-Newton as
theoretically expected.
Surprisingly, Gauss-Newton showed a better convergence curve than Newton in PDE-LDDMM based on the state equation.

\begin{table}

\begin{center}
\scriptsize
Benchmark methods, V-regularization, $\gamma = 0$
\\
\begin{tabular}{|c|c|c|c|c|c|c|c|c|}
\hline
Method & Opt. & $MSE_{rel}$ & $\Vert g\Vert_{\infty,{rel}}$ & $\max(J(\phi^v))$ & $\min(J(\phi^v))$ & PCG iter & $time_{GPU}(s)$\\
\hline
St. Mang & GN & 18.29 $\pm$ 2.83 & 0.07 $\pm$ 0.05 & 3.70 $\pm$ 0.51 & 0.16 $\pm$ 0.05 & 43.73 $\pm$ 10.64 & 3104.41 $\pm$ 695.22 \\
NSt. Mang & GN & 21.11 $\pm$ 5.18 & 0.20 $\pm$ 0.11 & 3.65 $\pm$ 0.85 & 0.15 $\pm$ 0.04 & 26.00 $\pm$ 15.21 & 5782.45 $\pm$ 1947.92 (*) \\
\hline
\end{tabular}
\begin{tabular}{|c|c|c|c|c|c|c|c|}
\hline
Method & Opt. & $MSE_{rel}$ & $\Vert g\Vert_{\infty,{rel}}$ & $\max(J(\phi^v))$ & $\min(J(\phi^v))$ & $time_{GPU}(s)$\\
\hline
Zhang & GD & 20.48 $\pm$ 1.84 & 0.04 $\pm$ 0.09 & 2.52 $\pm$ 0.22 & 0.19 $\pm$ 0.03 & 97.40 $\pm$ 10.06 & 4683.22 $\pm$ 1857.32 \\
\hline
BL Zhang, 64 & GD & 20.79 $\pm$ 2.03 & 0.01 $\pm$ 0.00 & 2.72 $\pm$ 0.31 & 0.18 $\pm$ 0.03 & 97.87 $\pm$ 5.73 & 1086.28 $\pm$ 44.60 \\
BL Zhang, 56 & GD & 20.80 $\pm$ 2.04 & 0.01 $\pm$ 0.00 & 2.72 $\pm$ 0.30 & 0.18 $\pm$ 0.03 & 97.67 $\pm$ 6.28 & 747.20 $\pm$ 32.76 \\
BL Zhang, 48 & GD & 20.87 $\pm$ 2.04 & 0.01 $\pm$ 0.00 & 2.74 $\pm$ 0.30 & 0.19 $\pm$ 0.03 & 98.00 $\pm$ 5.49 & 543.06 $\pm$ 28.55 \\
BL Zhang, 40 & GD & 21.00 $\pm$ 2.01 & 0.01 $\pm$ 0.00 & 2.71 $\pm$ 0.27 & 0.19 $\pm$ 0.03 & 98.87 $\pm$ 4.12 & 535.07 $\pm$ 25.87 \\
BL Zhang, 32 & GD & 21.33 $\pm$ 2.03 & 0.00 $\pm$ 0.00 & 2.64 $\pm$ 0.31 & 0.20 $\pm$ 0.03 & 98.60 $\pm$ 5.42 & 588.65 $\pm$ 24.56 \\
BL Zhang, 16 & GD & 25.77 $\pm$ 2.60 & 0.00 $\pm$ 0.00 & 2.40 $\pm$ 0.28 & 0.23 $\pm$ 0.04 & 100.00 $\pm$ 0.00 & 604.43 $\pm$ 25.21 \\
\hline
\end{tabular}
% \end{center}
% \caption{ Quantitative results of interest for the evaluation of the benchmark methods. 
% Recall that the implementation of NSt. Mang method was performed partially in the CPU.
% In Opt. column, GN stands for Gauss-Newton Krylov optimization and GD stands for gradient-descent optimization.}
% \label{table:QuantitativeResultsBenchmark}
% \end{table}
% 
% \begin{table}
% \begin{center}
% \scriptsize
\\
\vspace{0.1 cm}
St. BL PDE-LDDMM, state equation, $V$-regularization, $\gamma = 0$
\\
\begin{tabular}{|c|c|c|c|c|c|c|c|}
\hline
BL size & Opt. & $MSE_{rel}$ & $\Vert g\Vert_{\infty,{rel}}$ & $\max(J(\phi^v))$ & $\min(J(\phi^v))$ & PCG iter & $time_{GPU}$ (s) \\
\hline
64 & GN & 17.15 $\pm$ 1.58 & 0.02 $\pm$ 0.01 & 3.15 $\pm$ 0.37 & 0.12 $\pm$ 0.07 & 50.00 $\pm$ 0.00 & 423.73 $\pm$ 1.85 \\ 
56 & GN & 17.24 $\pm$ 1.59 & 0.02 $\pm$ 0.01 & 3.16 $\pm$ 0.37 & 0.12 $\pm$ 0.07 & 49.53 $\pm$ 0.52 & 381.33 $\pm$ 2.92 \\ 
48 & GN & 17.41 $\pm$ 1.60 & 0.02 $\pm$ 0.01 & 3.14 $\pm$ 0.36 & 0.11 $\pm$ 0.07 & 49.00 $\pm$ 0.00 & 372.71 $\pm$ 2.12 \\ 
40 & GN & 17.77 $\pm$ 1.63 & 0.02 $\pm$ 0.01 & 3.16 $\pm$ 0.38 & 0.11 $\pm$ 0.07 & 47.93 $\pm$ 0.96 & 364.03 $\pm$ 4.41 \\ 
32 & GN & 18.53 $\pm$ 1.71 & 0.02 $\pm$ 0.01 & 3.10 $\pm$ 0.35 & 0.11 $\pm$ 0.07 & 46.87 $\pm$ 0.35 & 358.78 $\pm$ 3.04 \\ 
16 & GN & 24.38 $\pm$ 2.34 & 0.02 $\pm$ 0.01 & 3.32 $\pm$ 0.73 & 0.17 $\pm$ 0.06 & 43.20 $\pm$ 1.08 & 335.40 $\pm$ 9.95 \\ 
\hline
\end{tabular}
\begin{tabular}{|c|c|c|c|c|c|c|c|}
\hline
BL size & Opt. & $MSE_{rel}$ & $\Vert g\Vert_{\infty,{rel}}$ & $\max(J(\phi^v))$ & $\min(J(\phi^v))$ & PCG iter & $time_{GPU}$ (s) \\
\hline
64 & N & 17.53 $\pm$ 1.60 & 0.03 $\pm$ 0.01 & 3.08 $\pm$ 0.28 & 0.11 $\pm$ 0.06 & 49.53 $\pm$ 0.64 & 853.19 $\pm$ 27.83 \\ 
56 & N & 17.62 $\pm$ 1.62 & 0.03 $\pm$ 0.01 & 3.06 $\pm$ 0.31 & 0.11 $\pm$ 0.06 & 48.73 $\pm$ 0.59 & 710.48 $\pm$ 5.52 \\ 
48 & N & 17.80 $\pm$ 1.63 & 0.03 $\pm$ 0.01 & 3.08 $\pm$ 0.33 & 0.11 $\pm$ 0.06 & 48.13 $\pm$ 0.83 & 644.01 $\pm$ 18.06 \\ 
40 & N & 18.16 $\pm$ 1.69 & 0.03 $\pm$ 0.01 & 3.13 $\pm$ 0.38 & 0.11 $\pm$ 0.07 & 47.00 $\pm$ 0.38 & 598.04 $\pm$ 27.39 \\ 
32 & N & 18.94 $\pm$ 1.75 & 0.03 $\pm$ 0.01 & 3.11 $\pm$ 0.34 & 0.11 $\pm$ 0.07 & 46.07 $\pm$ 0.96 & 554.64 $\pm$ 7.10 \\ 
16 & N & 24.95 $\pm$ 2.42 & 0.02 $\pm$ 0.01 & 3.31 $\pm$ 0.81 & 0.16 $\pm$ 0.06 & 41.73 $\pm$ 1.03 & 480.10 $\pm$ 8.11 \\ 
\hline
\end{tabular}
\\
\vspace{0.1 cm}
St. BL PDE-LDDMM, deformation state equation, $V$-regularization, $\gamma = 0$
\\
\begin{tabular}{|c|c|c|c|c|c|c|c|}
\hline
BL size & Opt. & $MSE_{rel}$ & $\Vert g\Vert_{\infty,{rel}}$ & $\max(J(\phi^v))$ & $\min(J(\phi^v))$ & PCG iter & $time_{GPU}$ (s) \\
\hline
64 & GN & 16.04 $\pm$ 1.53 & 0.04 $\pm$ 0.01 & 3.88 $\pm$ 0.42 & 0.11 $\pm$ 0.04 & 49.47 $\pm$ 1.46 & 586.51 $\pm$ 9.33 \\ 
56 & GN & 16.09 $\pm$ 1.52 & 0.03 $\pm$ 0.01 & 3.86 $\pm$ 0.41 & 0.11 $\pm$ 0.04 & 49.53 $\pm$ 0.52 & 502.26 $\pm$ 10.10 \\ 
48 & GN & 16.25 $\pm$ 1.55 & 0.03 $\pm$ 0.01 & 3.84 $\pm$ 0.41 & 0.12 $\pm$ 0.05 & 49.00 $\pm$ 0.00 & 439.04 $\pm$ 9.63 \\ 
40 & GN & 16.57 $\pm$ 1.59 & 0.03 $\pm$ 0.01 & 3.78 $\pm$ 0.48 & 0.13 $\pm$ 0.05 & 48.00 $\pm$ 1.00 & 402.28 $\pm$ 8.00 \\ 
32 & GN & 17.32 $\pm$ 1.68 & 0.03 $\pm$ 0.01 & 3.76 $\pm$ 0.49 & 0.13 $\pm$ 0.05 & 46.80 $\pm$ 0.56 & 386.06 $\pm$ 10.88 \\ 
16 & GN & 23.34 $\pm$ 2.29 & 0.03 $\pm$ 0.01 & 3.30 $\pm$ 0.61 & 0.17 $\pm$ 0.05 & 42.07 $\pm$ 1.28 & 334.60 $\pm$ 10.85 \\ 
\hline
\end{tabular}
\begin{tabular}{|c|c|c|c|c|c|c|c|}
\hline
BL size & Opt. & $MSE_{rel}$ & $\Vert g\Vert_{\infty,{rel}}$ & $\max(J(\phi^v))$ & $\min(J(\phi^v))$ & PCG iter & $time_{GPU}$ (s) \\
\hline
64 & N & 14.75 $\pm$ 1.38 & 0.10 $\pm$ 0.03 & 4.89 $\pm$ 0.68 & 0.08 $\pm$ 0.04 & 49.53 $\pm$ 0.83 & 900.40 $\pm$ 12.62 \\ 
56 & N & 14.83 $\pm$ 1.38 & 0.11 $\pm$ 0.03 & 4.95 $\pm$ 0.72 & 0.08 $\pm$ 0.04 & 49.53 $\pm$ 0.83 & 752.90 $\pm$ 12.64 \\ 
48 & N & 15.02 $\pm$ 1.40 & 0.12 $\pm$ 0.03 & 5.06 $\pm$ 0.77 & 0.08 $\pm$ 0.04 & 49.93 $\pm$ 0.26 & 666.53 $\pm$ 9.22 \\ 
40 & N & 15.38 $\pm$ 1.43 & 0.11 $\pm$ 0.03 & 5.01 $\pm$ 0.74 & 0.08 $\pm$ 0.04 & 50.00 $\pm$ 0.00 & 593.95 $\pm$ 7.08 \\ 
32 & N & 16.19 $\pm$ 1.50 & 0.11 $\pm$ 0.03 & 4.92 $\pm$ 0.68 & 0.09 $\pm$ 0.04 & 50.00 $\pm$ 0.00 & 573.59 $\pm$ 8.87 \\ 
16 & N & 22.69 $\pm$ 2.26 & 0.08 $\pm$ 0.03 & 4.21 $\pm$ 0.77 & 0.11 $\pm$ 0.03 & 50.00 $\pm$ 0.00 & 514.33 $\pm$ 6.23 \\ 
\hline
\end{tabular}
\\
\vspace{0.1 cm}
NSt. BL PDE-LDDMM, deformation state equation, V-regularization, $\gamma = 0$
\\
\begin{tabular}{|c|c|c|c|c|c|c|c|}
\hline
BL size & Opt. & $MSE_{rel}$ & $\Vert g\Vert_{\infty,{rel}}$ & $\max(J(\phi^v))$ & $\min(J(\phi^v))$ & PCG iter & $time_{GPU}$ (s) \\
\hline
40 & GN & 14.93 $\pm$ 1.40 & 0.04 $\pm$ 0.01 & 4.87 $\pm$ 0.77 & 0.08 $\pm$ 0.04 & 48.40 $\pm$ 0.83 & 6427.13 $\pm$ 346.89 \\
32 & GN & 15.72 $\pm$ 1.52 & 0.04 $\pm$ 0.01 & 4.74 $\pm$ 0.77 & 0.09 $\pm$ 0.04 & 46.87 $\pm$ 0.35 & 5729.49 $\pm$ 306.90 \\
16 & GN & 21.94 $\pm$ 2.12 & 0.04 $\pm$ 0.01 & 3.96 $\pm$ 0.89 & 0.13 $\pm$ 0.04 & 42.47 $\pm$ 1.46 & 4729.42 $\pm$ 201.01 \\
\hline
40 & N & 14.83 $\pm$ 1.33 & 0.11 $\pm$ 0.04 & 6.91 $\pm$ 2.16 & 0.05 $\pm$ 0.03 & 49.80 $\pm$ 0.41 & 8098.13 $\pm$ 308.54 \\
32 & N & 15.59 $\pm$ 1.39 & 0.11 $\pm$ 0.03 & 6.23 $\pm$ 1.32 & 0.05 $\pm$ 0.03 & 50.00 $\pm$ 0.00 & 8609.06 $\pm$ 384.97 \\
16 & N & 22.19 $\pm$ 2.23 & 0.09 $\pm$ 0.03 & 4.89 $\pm$ 1.06 & 0.08 $\pm$ 0.03 & 49.60 $\pm$ 1.12 & 7167.19 $\pm$ 330.20 \\
\hline 
\end{tabular}

% BL PDE-LDDMM, state equation, $V$-regularization, $\gamma = 0$
% \\
% \begin{tabular}{|c|c|c|c|c|c|c|c|}
% \hline
% BL size & Opt. & $MSE_{rel}$ & $\Vert g\Vert_{\infty,{rel}}$ & $\max(J(\phi_{1}^v)^{-1})$ & $\min(J(\phi_{1}^v)^{-1})$ & PCG iter & $time_{GPU}$ (s) \\
% \hline
% $64 \times 64 \times 64$ & GN & & & & & & \\
% \hline
% $56 \times 56 \times 56$ & GN & & & & & & \\
% \hline
% $48 \times 48 \times 48$ & GN & & & & & & \\
% \hline
% $40 \times 40 \times 40$ & GN & & & & & & \\
% \hline
% $32 \times 32 \times 32$ & GN & & & & & & \\
% \hline
% $16 \times 16 \times 16$ & GN & & & & & & \\
% \hline
% \end{tabular}

\end{center}

\caption{ \small Quantitative results relevant for the evaluation of the benchmark and the proposed methods.
% Results obtained with $V$-regularization.
% Mean and standard deviation of the image similarity errors after registration, 
% the relative gradient magnitude, 
% the Jacobian determinant extrema associated with the transformation $\phi$, 
% the number of PCG iterations, and the computation time in the GPU.
% % For comparison purposes, results from the bencmark methods Stationary Mang and 
% % Non-Stationary Mang~\cite{Mang_15}, and spatial and band-limited FLASH-LDDMM~\cite{Zhang_15} 
% % are included.
In Opt. column, GD stands for gradient-descent optimization, 
and N and GN stand for Newton- and Gauss-Newton- Krylov optimization, respectively.
(*) Recall that the implementation of NSt. Mang method was performed partially in the CPU.
}
\label{table:QuantitativeResultsV}
\end{table}

\subsection{Qualitative results}

\begin{figure*} [!t]
\centering
\scriptsize
\begin{tabular}{cc}
\includegraphics[angle = 0, width=2.0 cm, height = 1.75 cm]{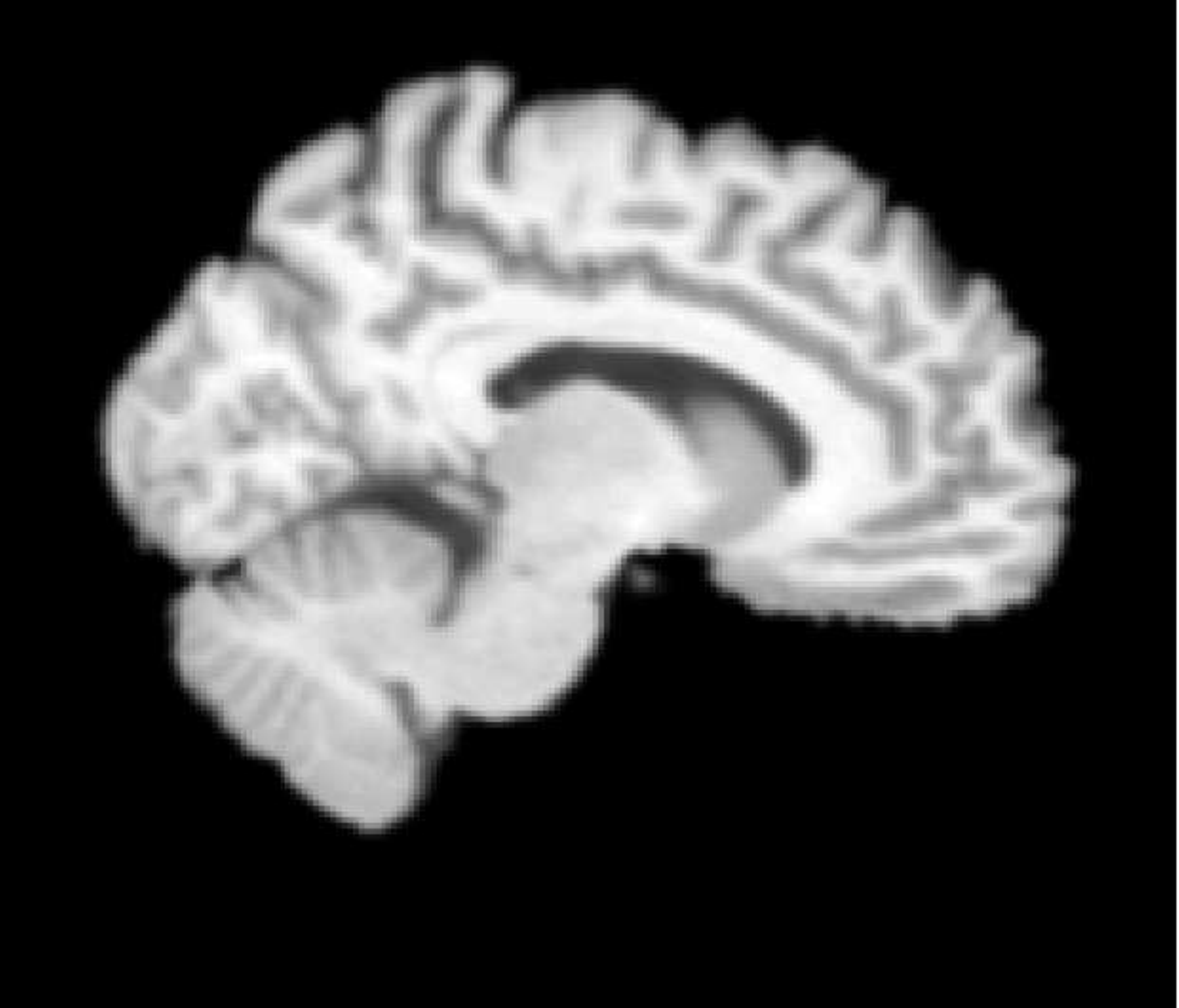} 
&
\includegraphics[angle = 0, width=2.0 cm, height = 1.75 cm]{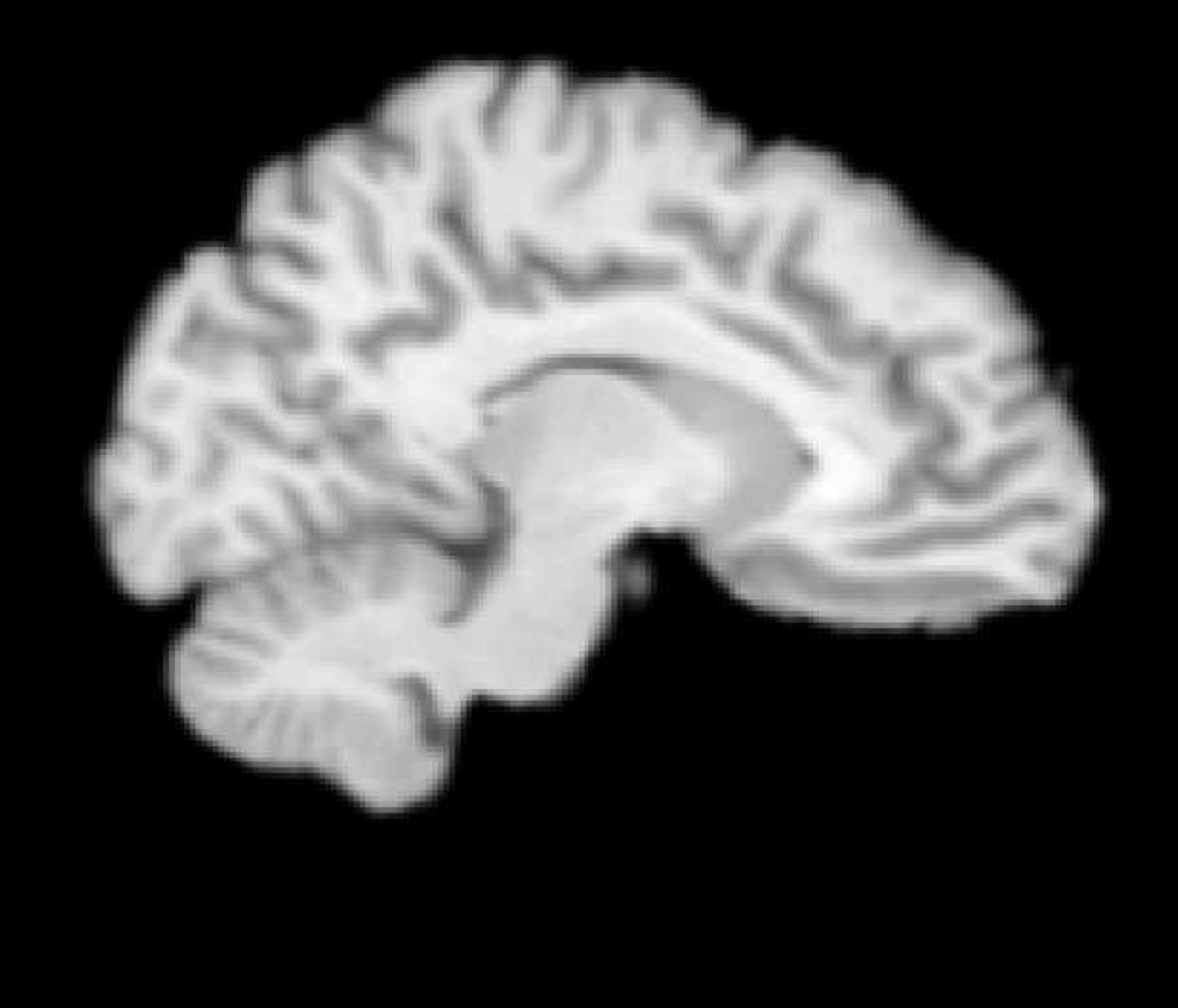} \\
source & target \\
\end{tabular}
\begin{tabular}{cccc} 
\includegraphics[angle = 0, width=2.0 cm, height = 1.75 cm]{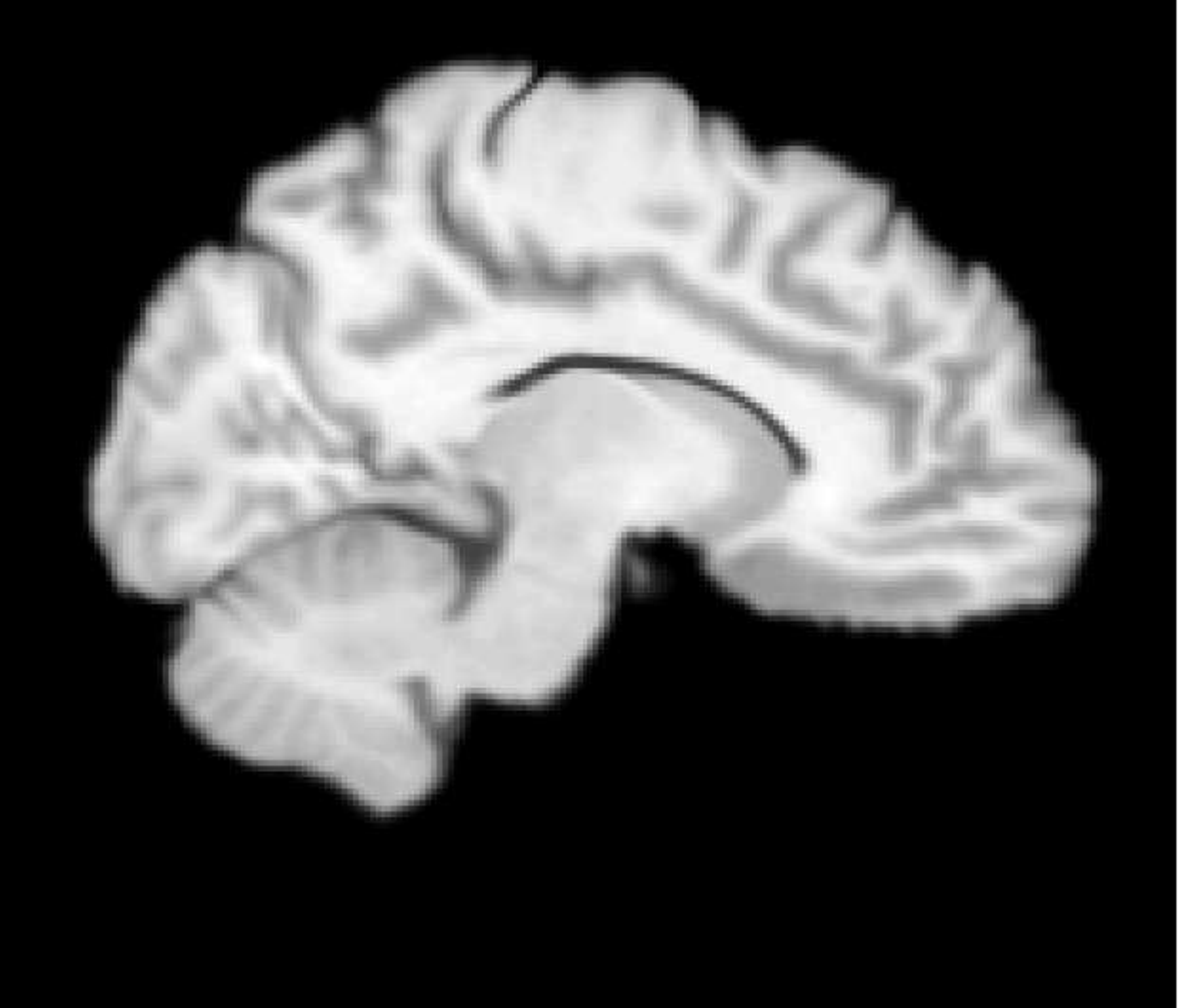} 
&
\includegraphics[angle = 0, width=2.0 cm, height = 1.75 cm]{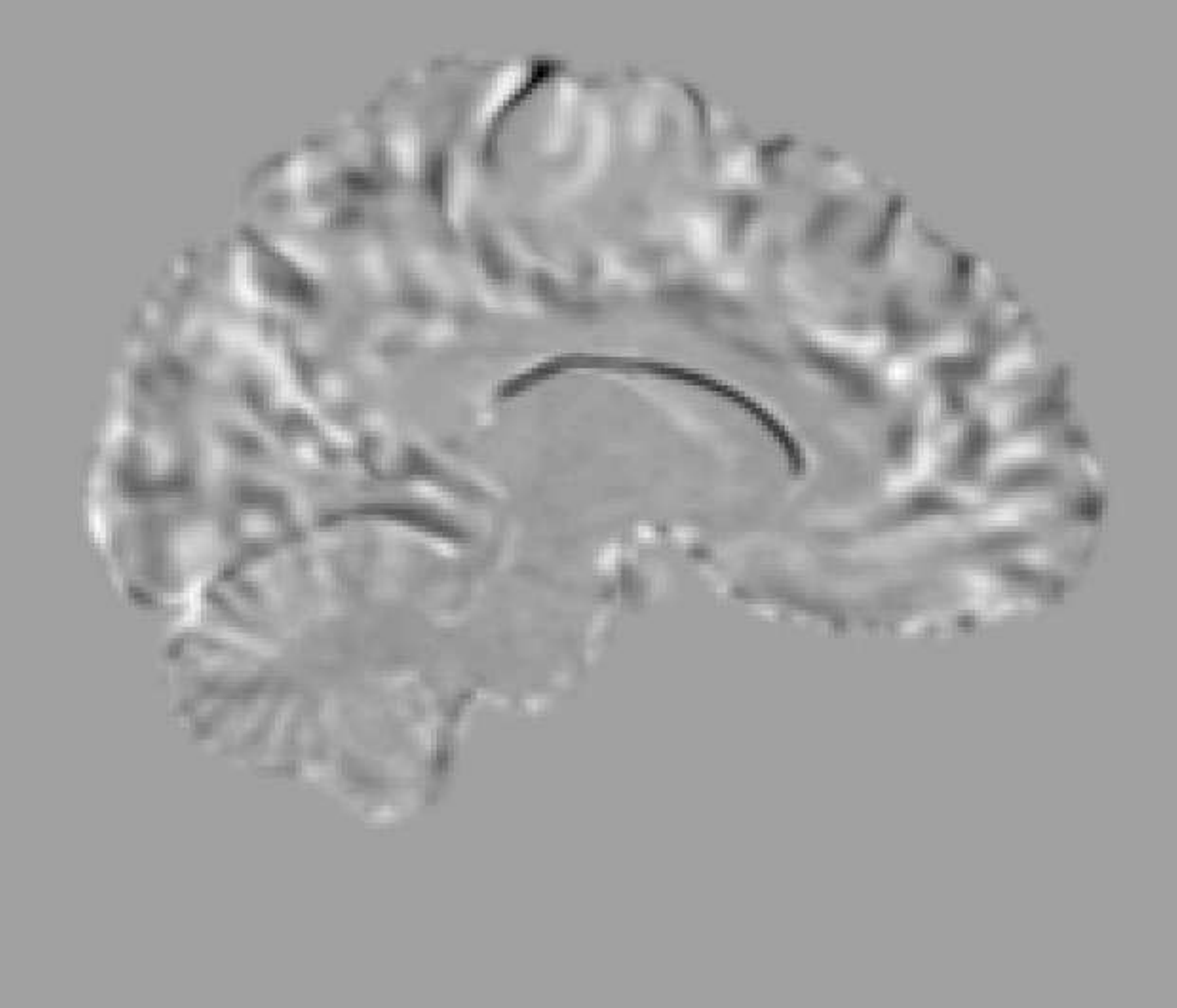} 
&
\includegraphics[angle = 0, width=2.0 cm, height = 1.75 cm]{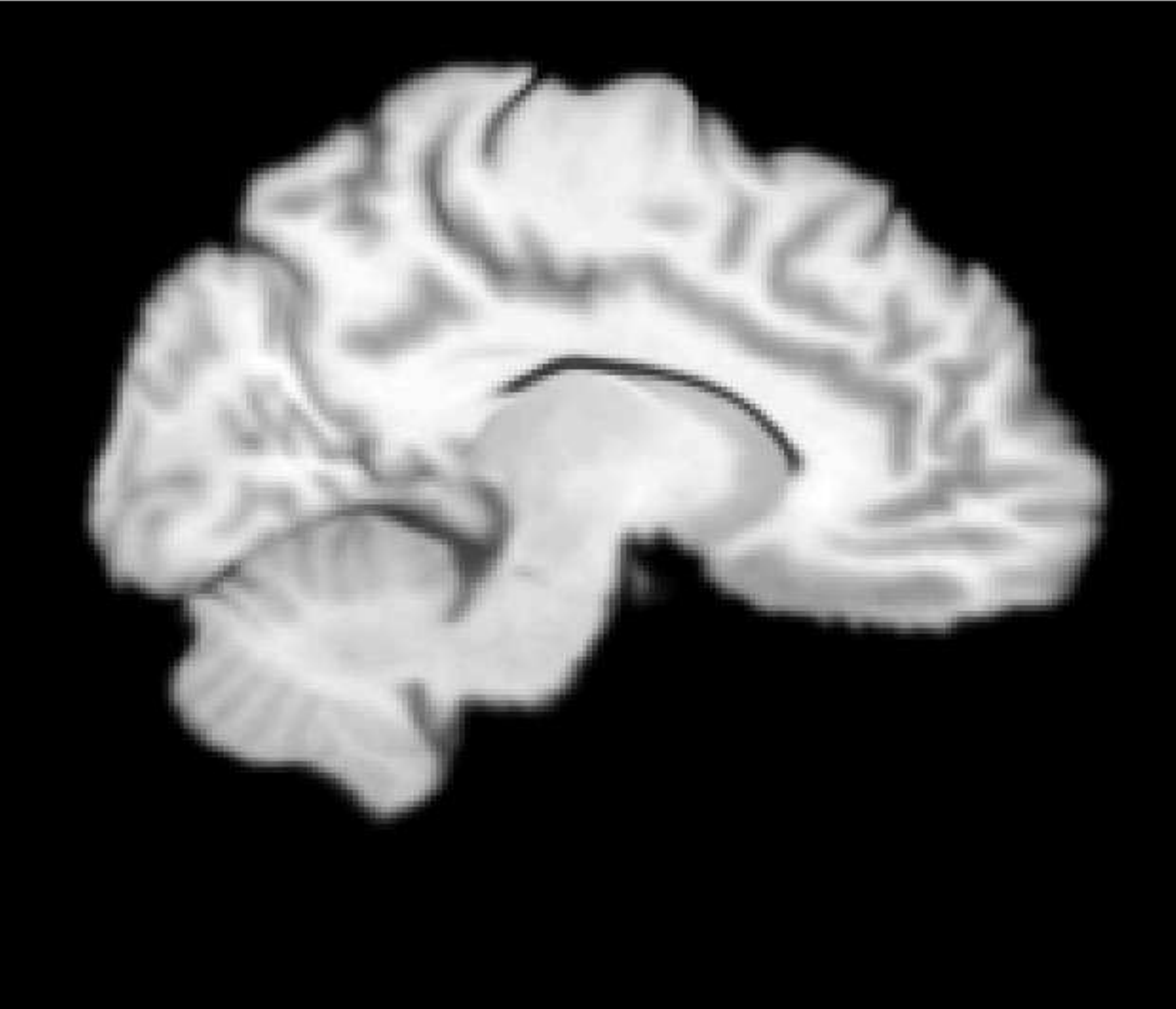} 
&
\includegraphics[angle = 0, width=2.0 cm, height = 1.75 cm]{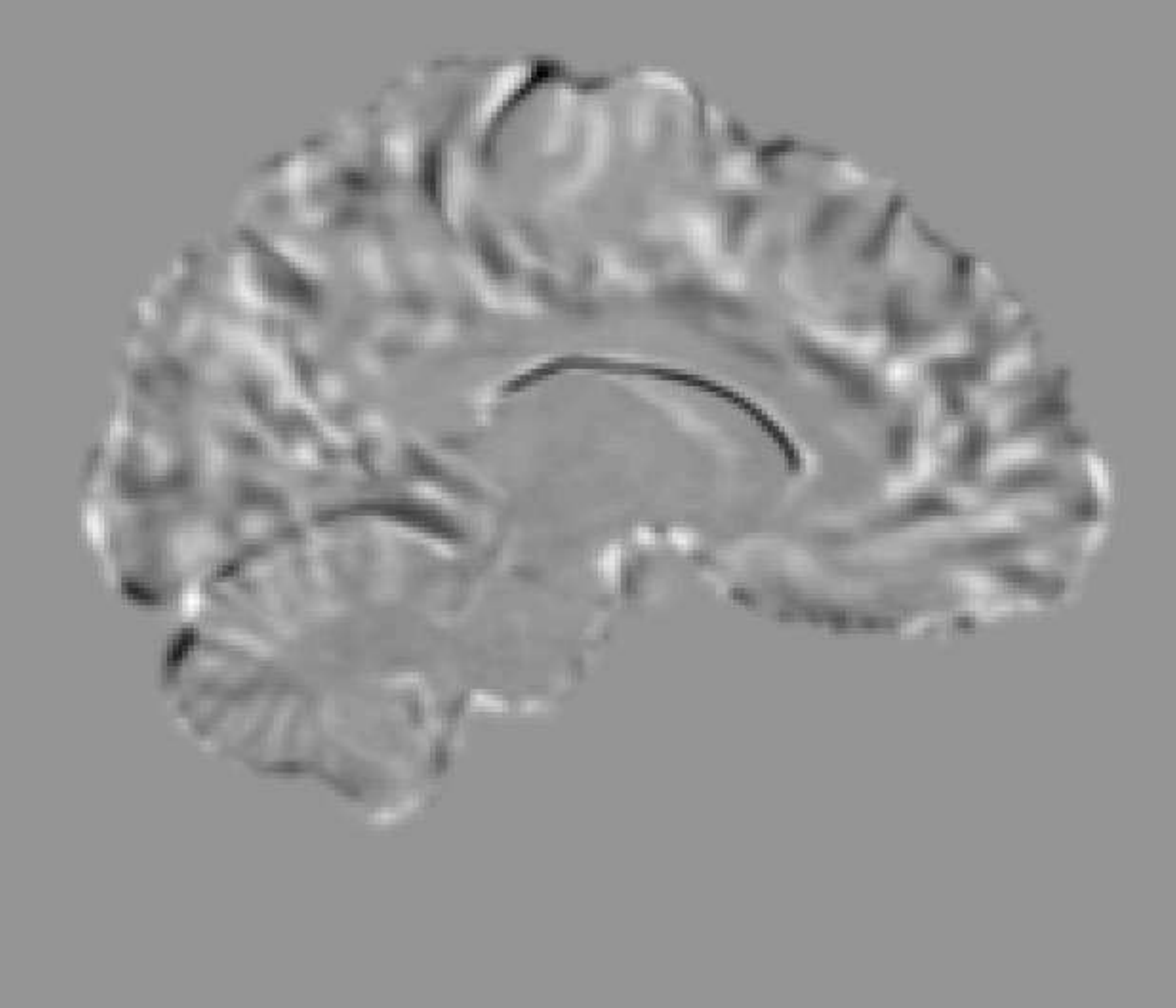} 
\\
St. Mang & & NSt. Mang & 
\\
\includegraphics[angle = 0, width=2.0 cm, height = 1.75 cm]{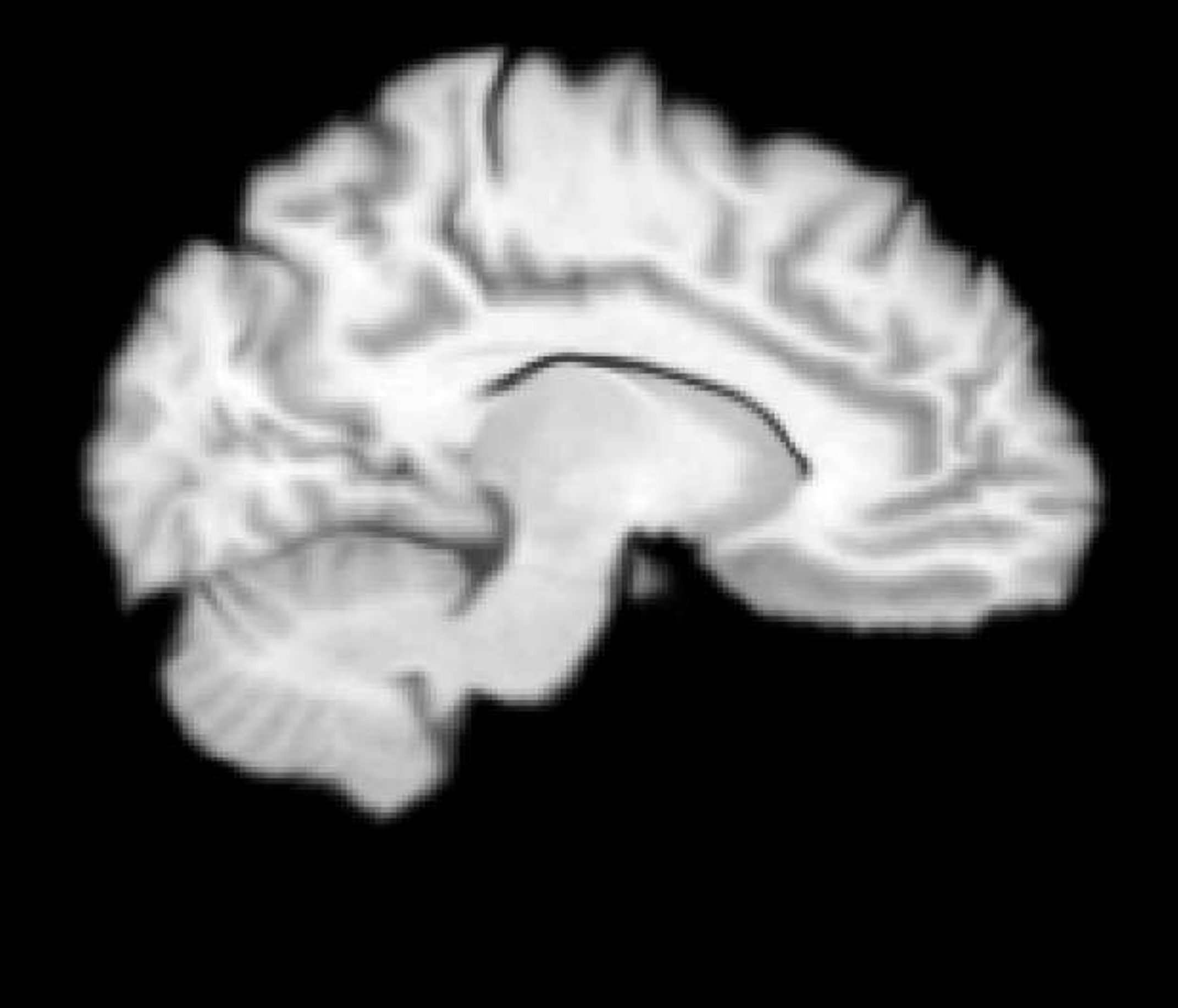} 
&
\includegraphics[angle = 0, width=2.0 cm, height = 1.75 cm]{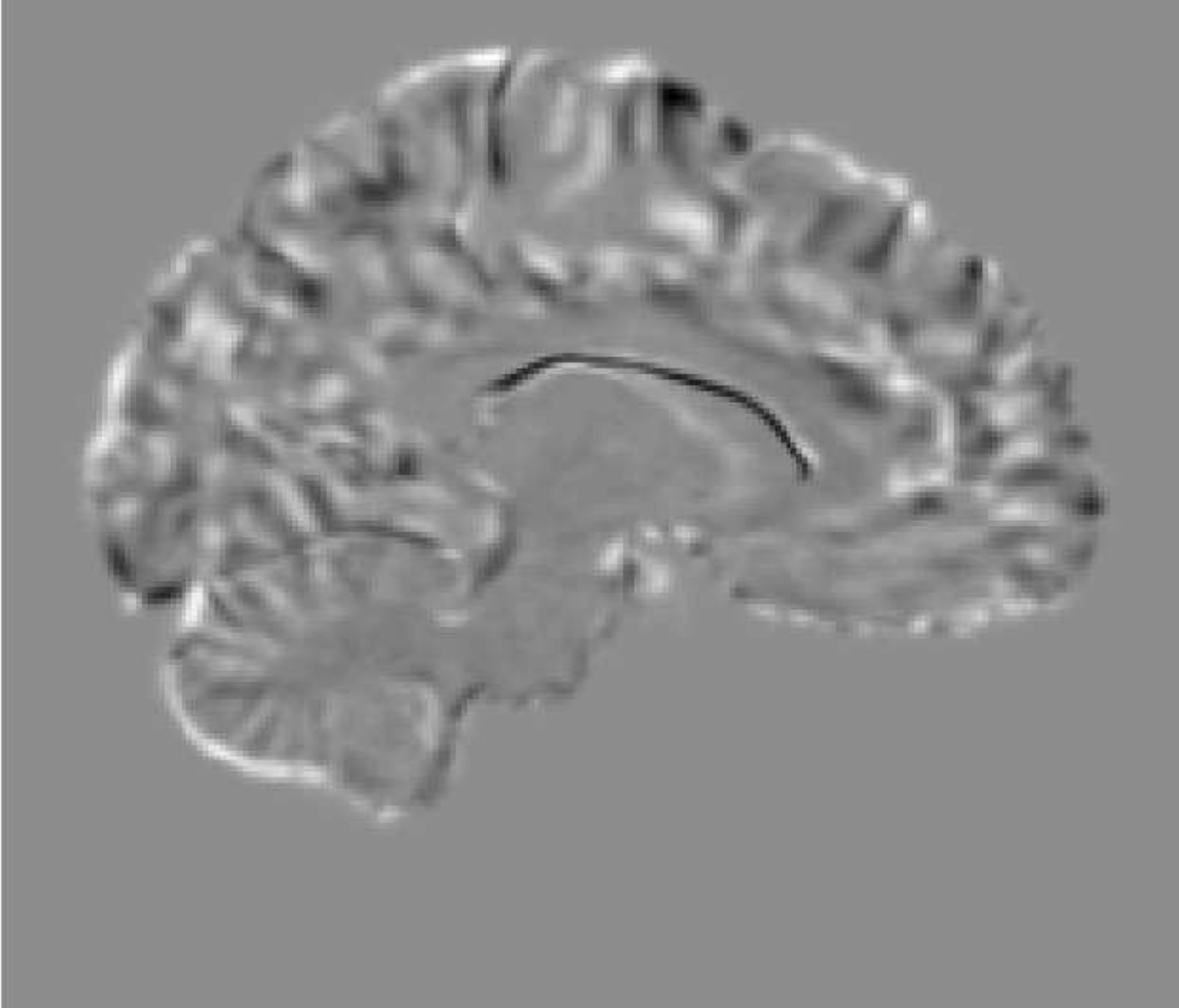} 
&
\includegraphics[angle = 0, width=2.0 cm, height = 1.75 cm]{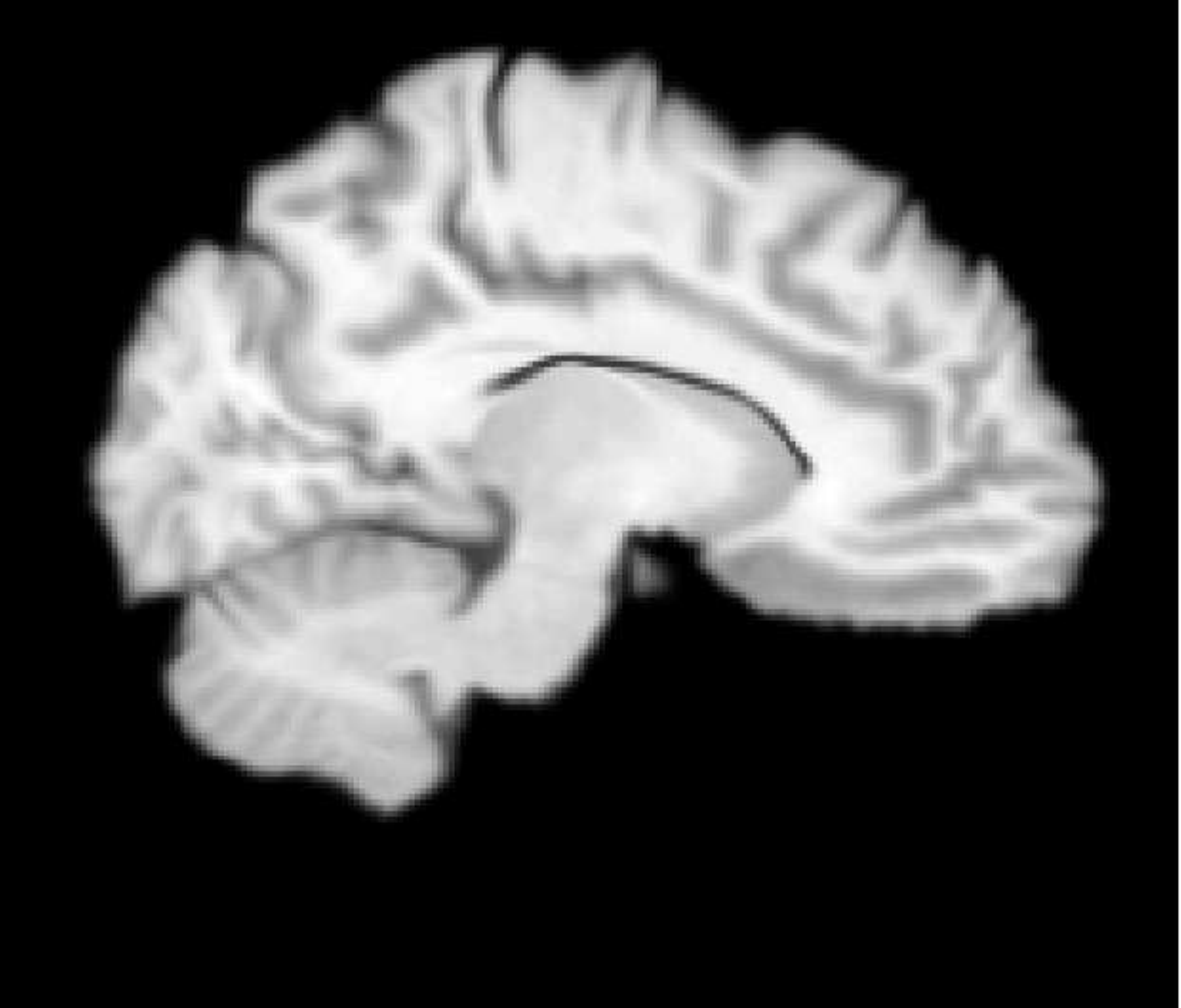} 
&
\includegraphics[angle = 0, width=2.0 cm, height = 1.75 cm]{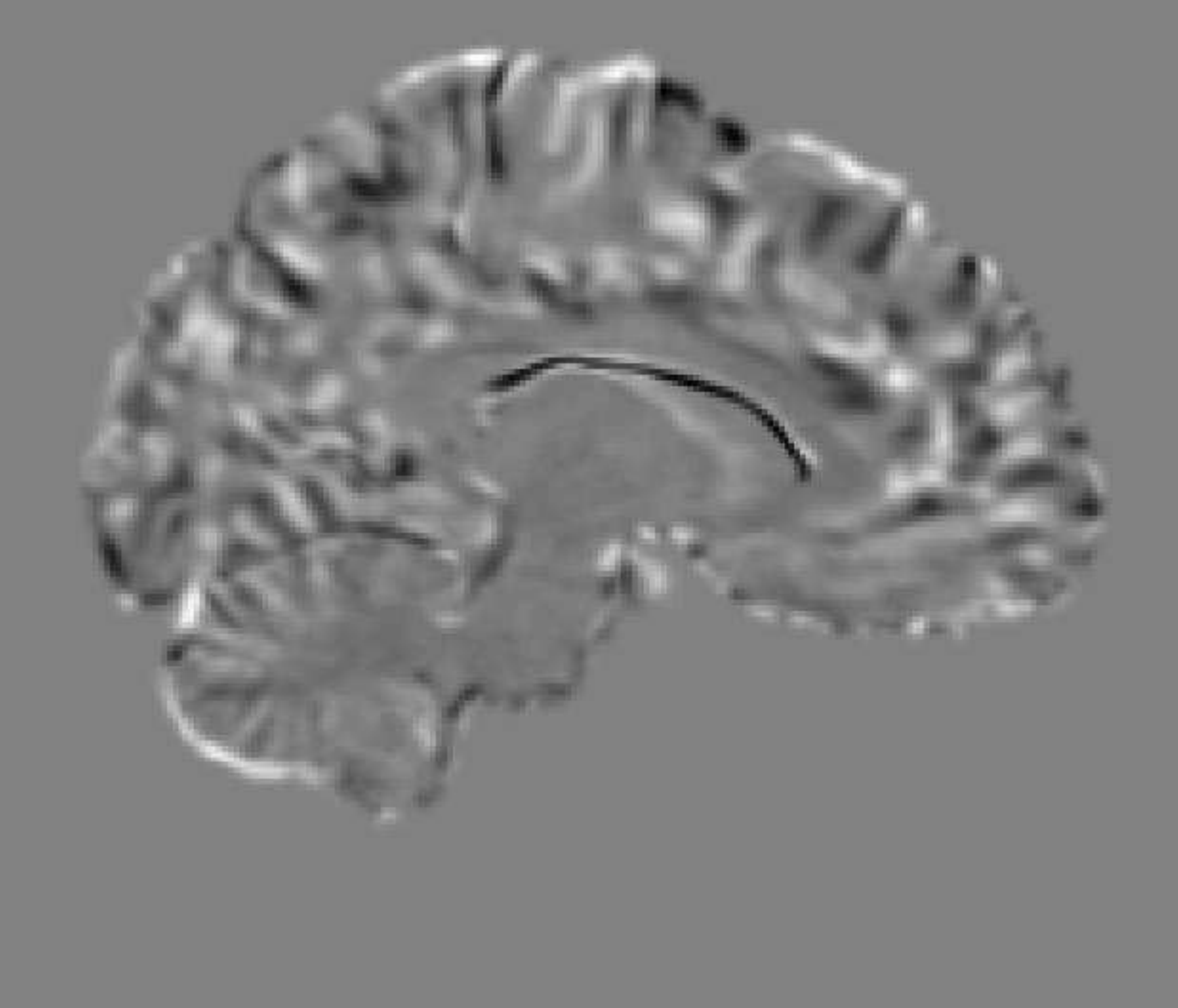} 
\\
Zhang & & BL Zhang & 
\\
\includegraphics[angle = 0, width=2.0 cm, height = 1.75 cm]{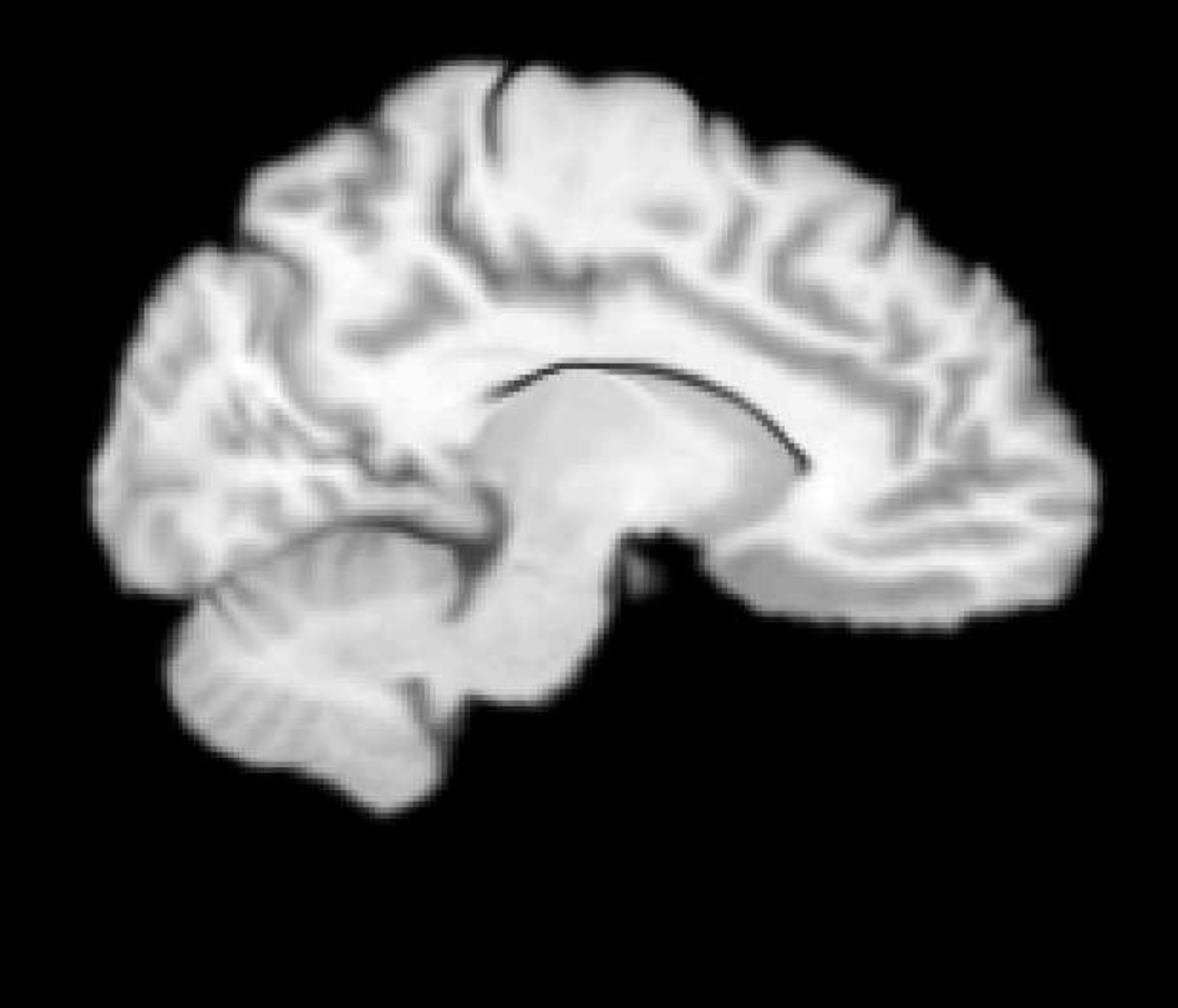} 
&
\includegraphics[angle = 0, width=2.0 cm, height = 1.75 cm]{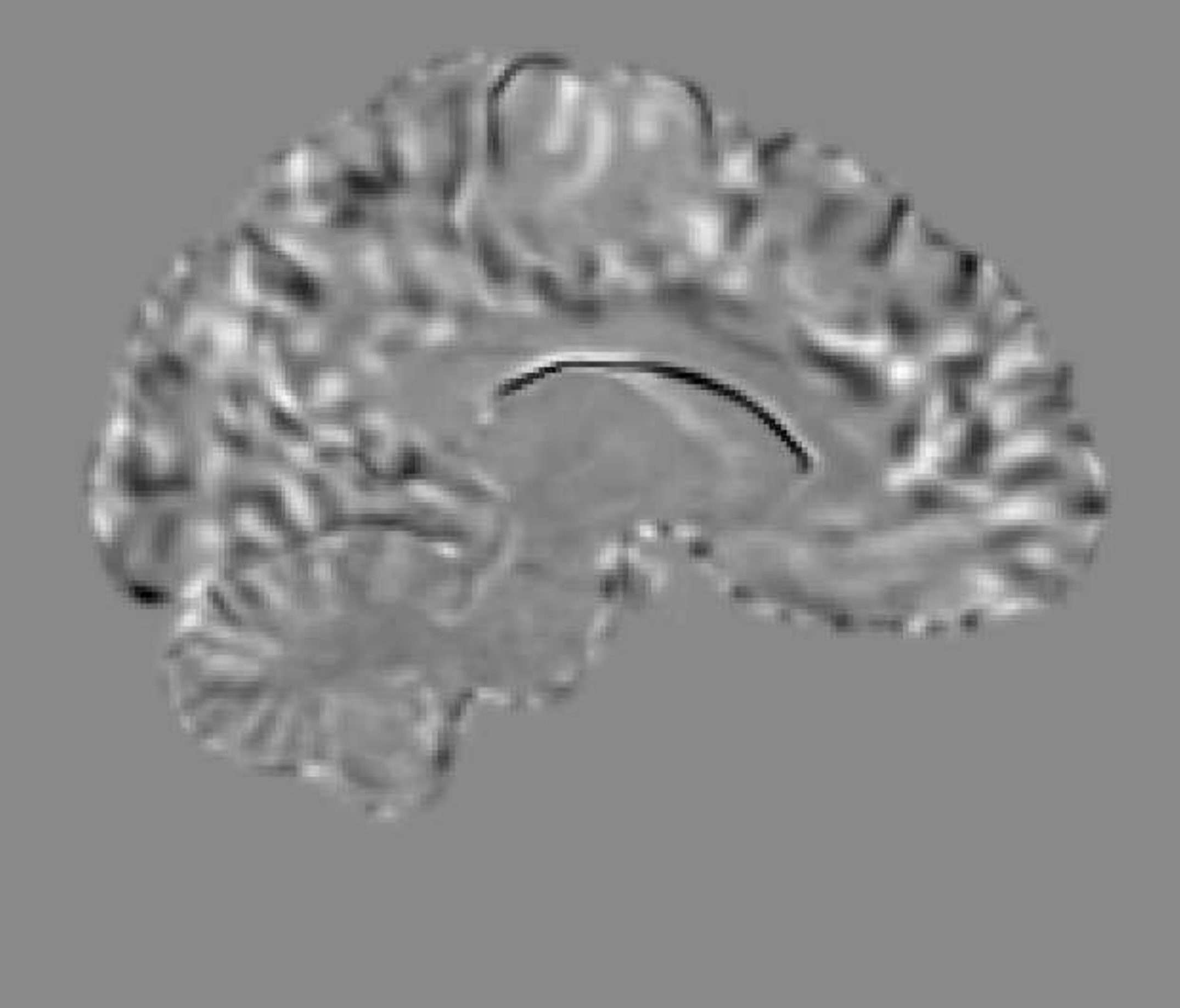} 
&
\includegraphics[angle = 0, width=2.0 cm, height = 1.75 cm]{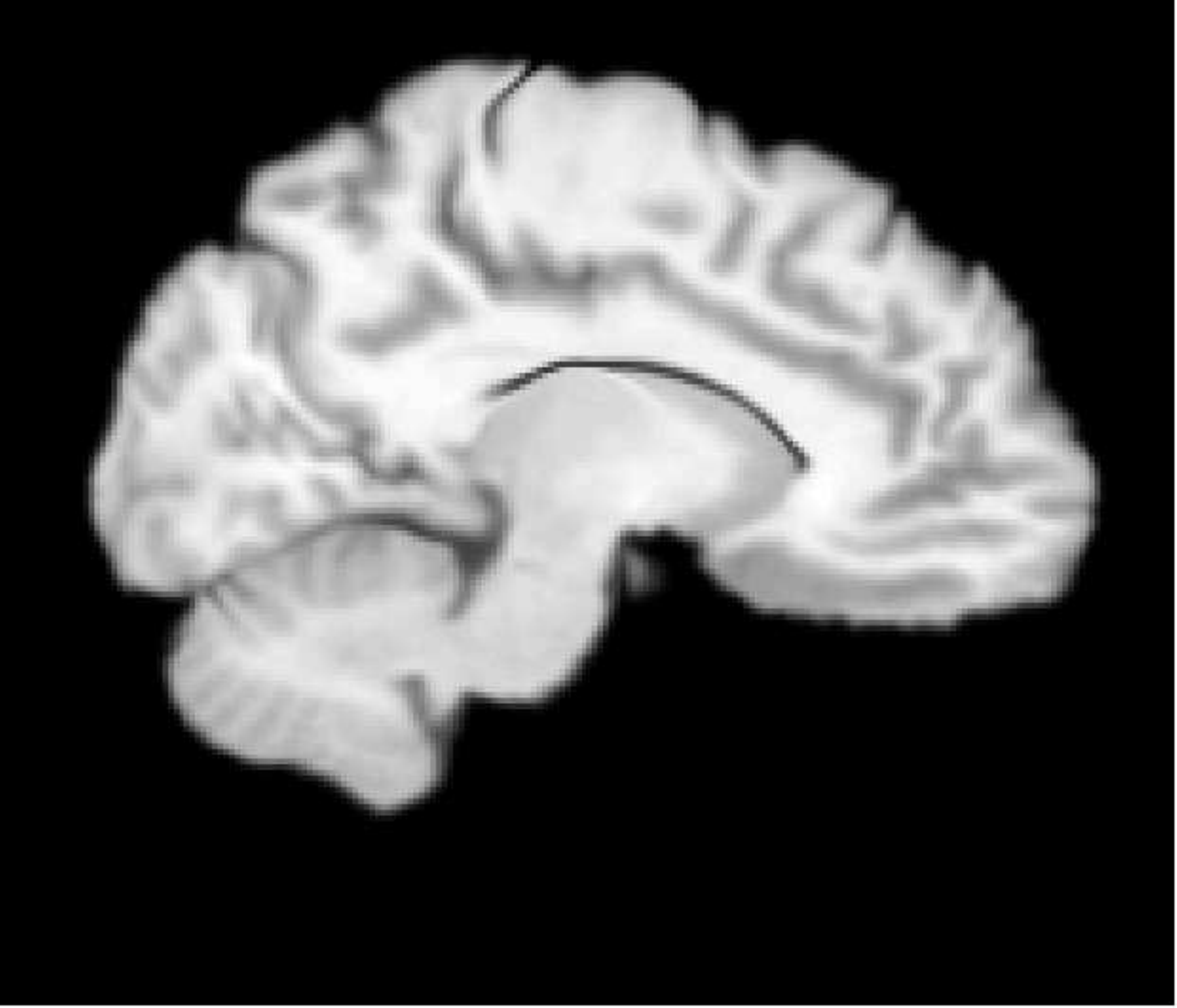} 
&
\includegraphics[angle = 0, width=2.0 cm, height = 1.75 cm]{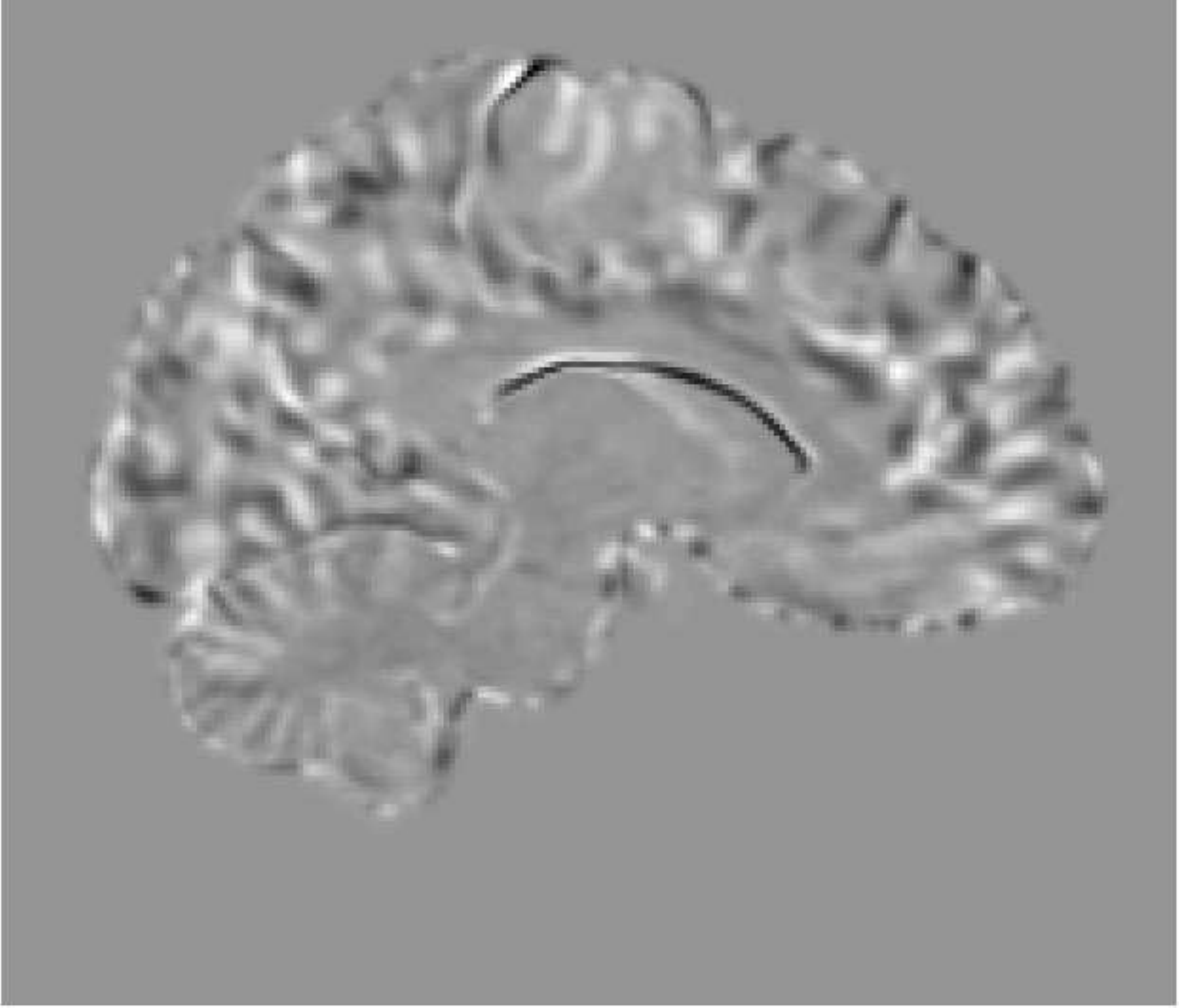} 
\\
\begin{tabular}{c} St. BL PDE-LDDMM \\ state equation, GN \end{tabular} & & \begin{tabular}{c} St. BL PDE-LDDMM \\ state equation, N \end{tabular} 
\\
\includegraphics[angle = 0, width=2.0 cm, height = 1.75 cm]{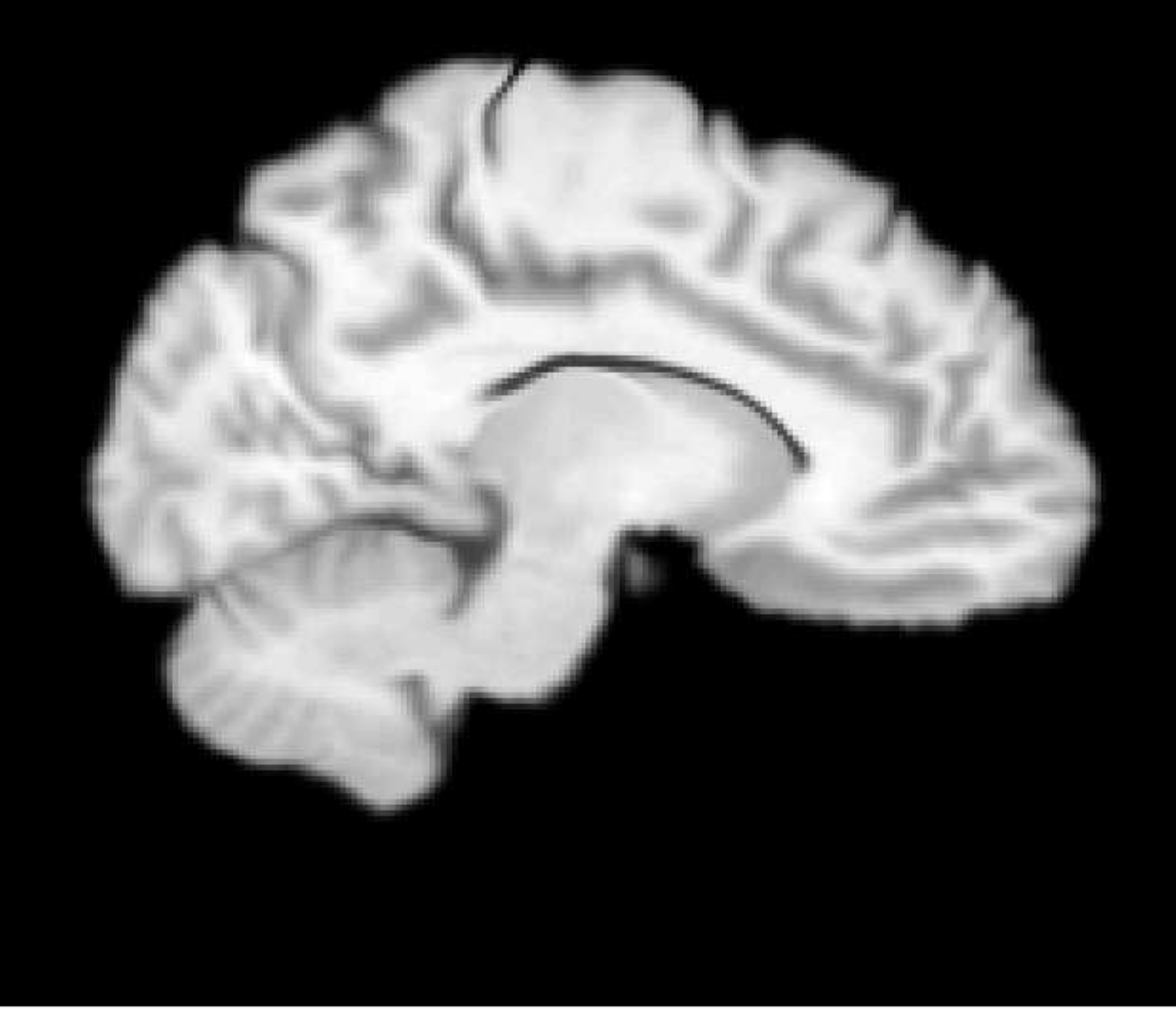} 
&
\includegraphics[angle = 0, width=2.0 cm, height = 1.75 cm]{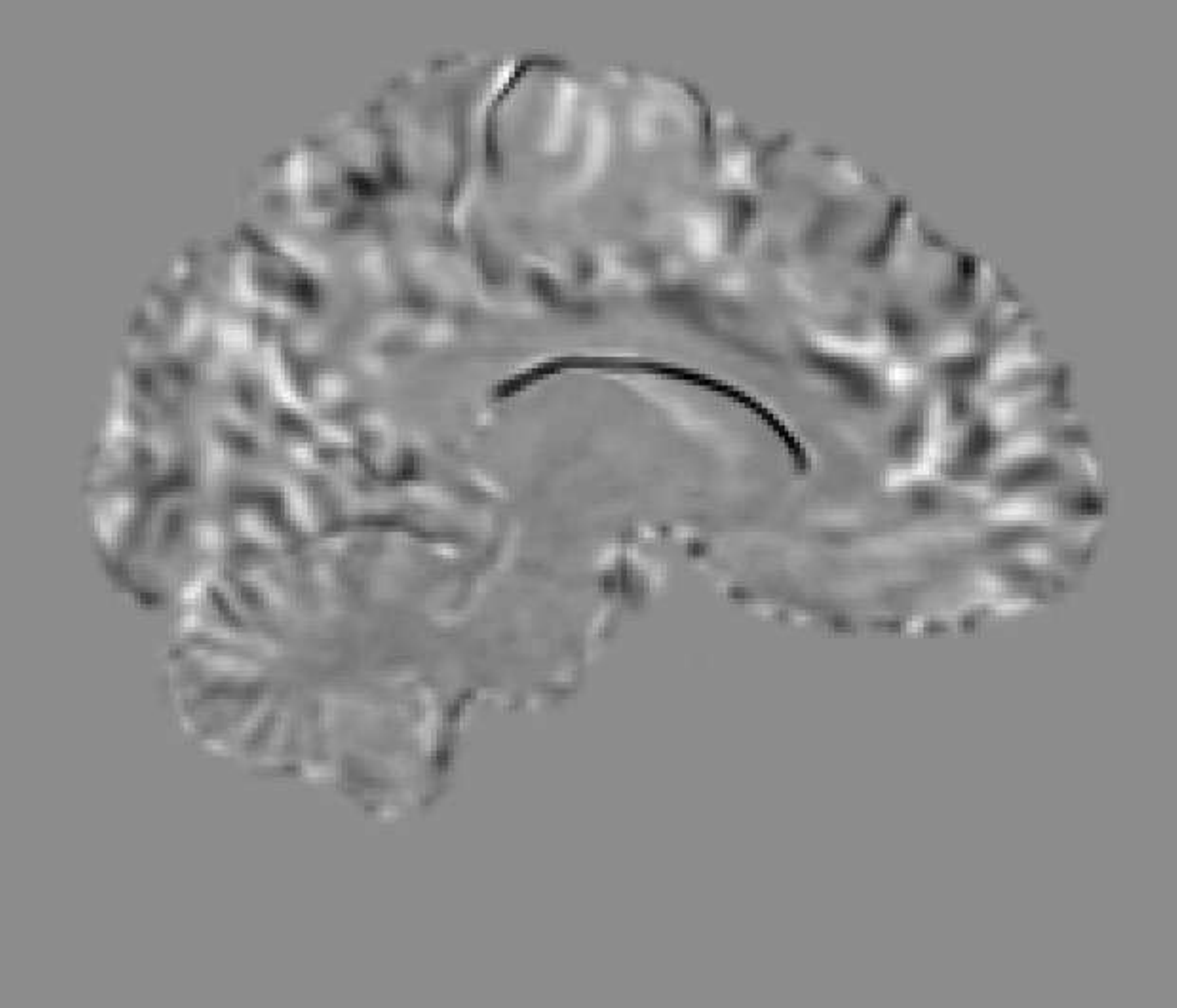} 
&
\includegraphics[angle = 0, width=2.0 cm, height = 1.75 cm]{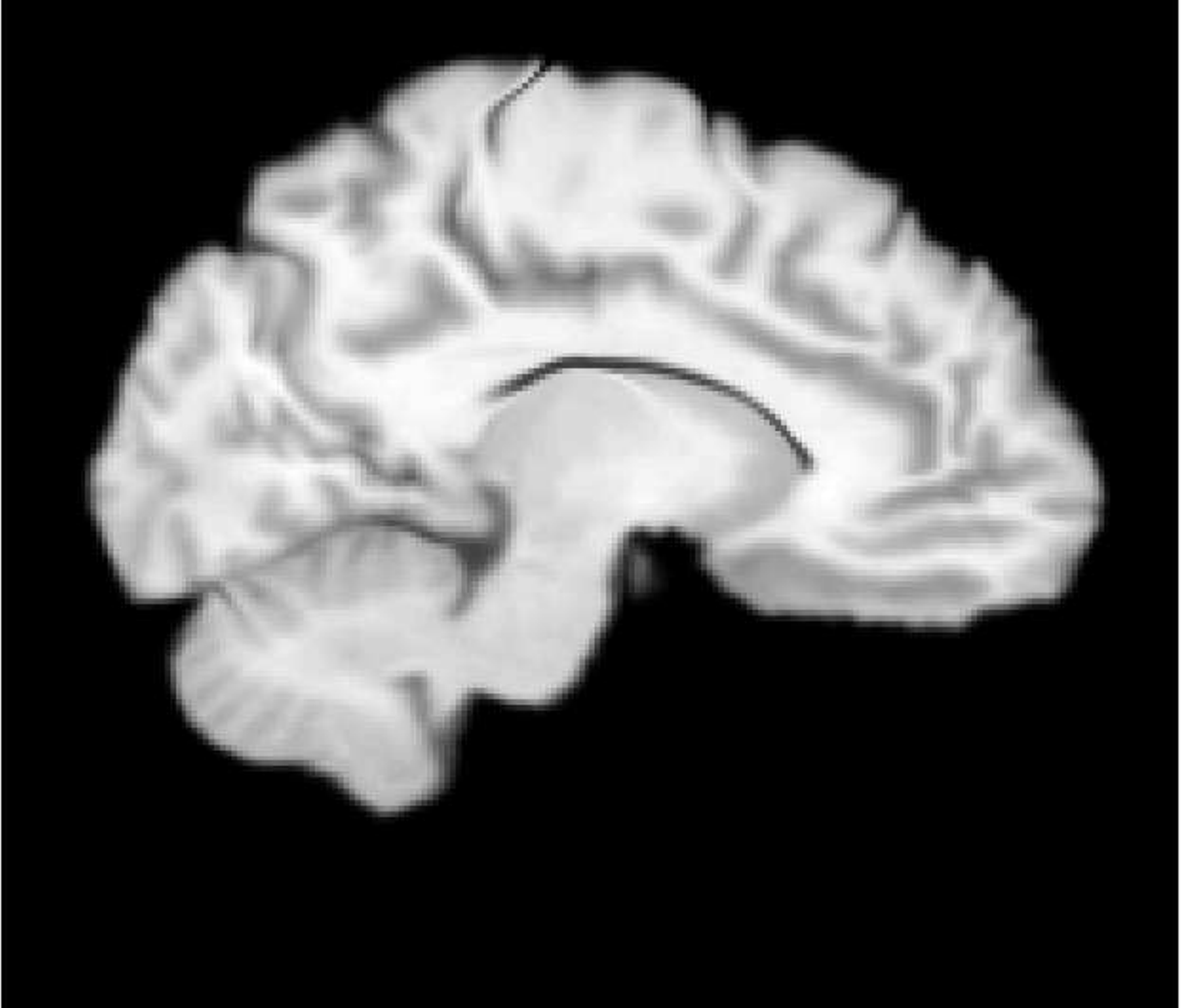} 
&
\includegraphics[angle = 0, width=2.0 cm, height = 1.75 cm]{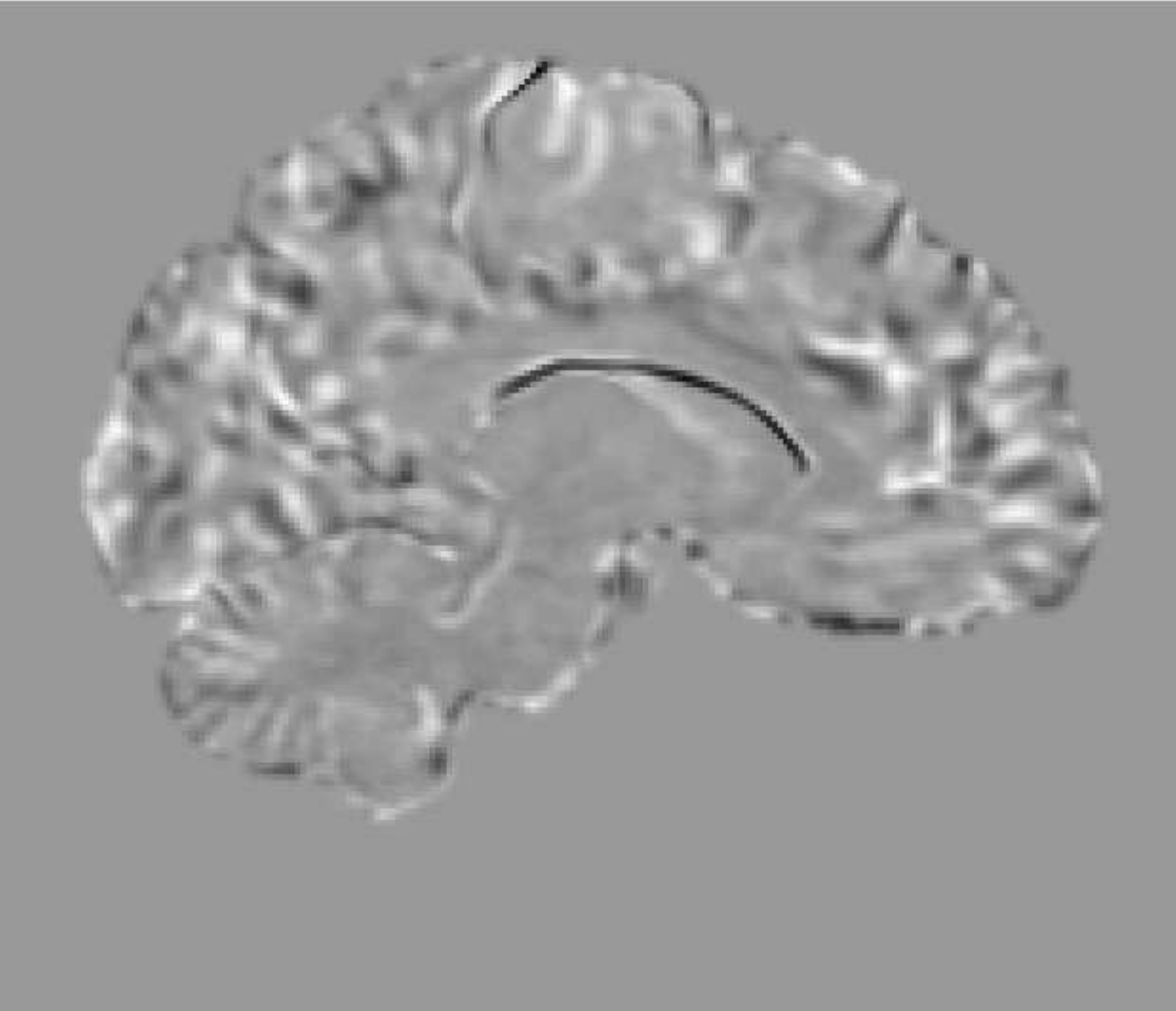} 
\\
\begin{tabular}{c} St. BL PDE-LDDMM \\ def. equation, GN \end{tabular} & & \begin{tabular}{c} St. BL PDE-LDDMM \\ def. equation, N \end{tabular} & 
\\
\\
\includegraphics[angle = 0, width=2.0 cm, height = 1.75 cm]{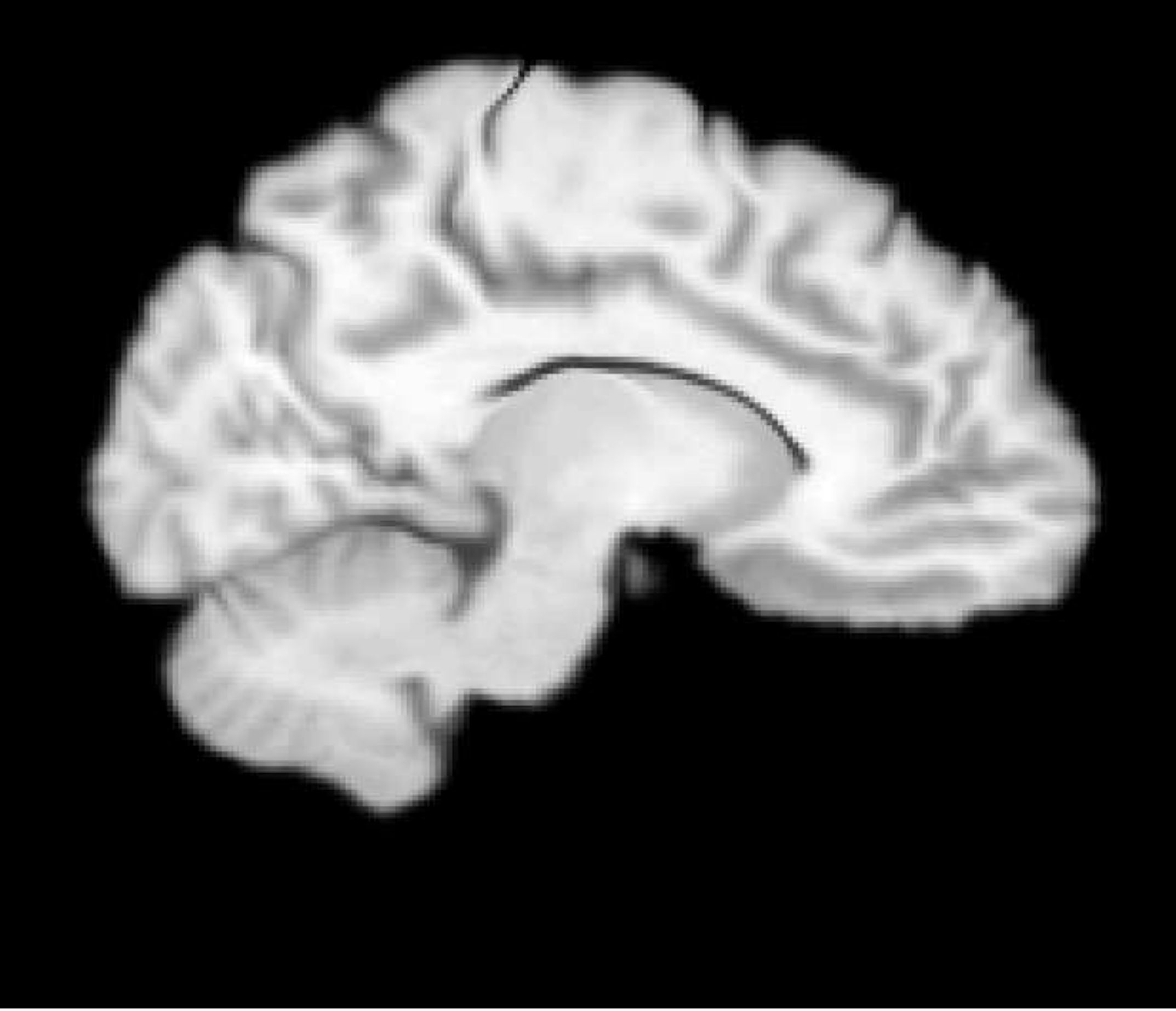} 
&
\includegraphics[angle = 0, width=2.0 cm, height = 1.75 cm]{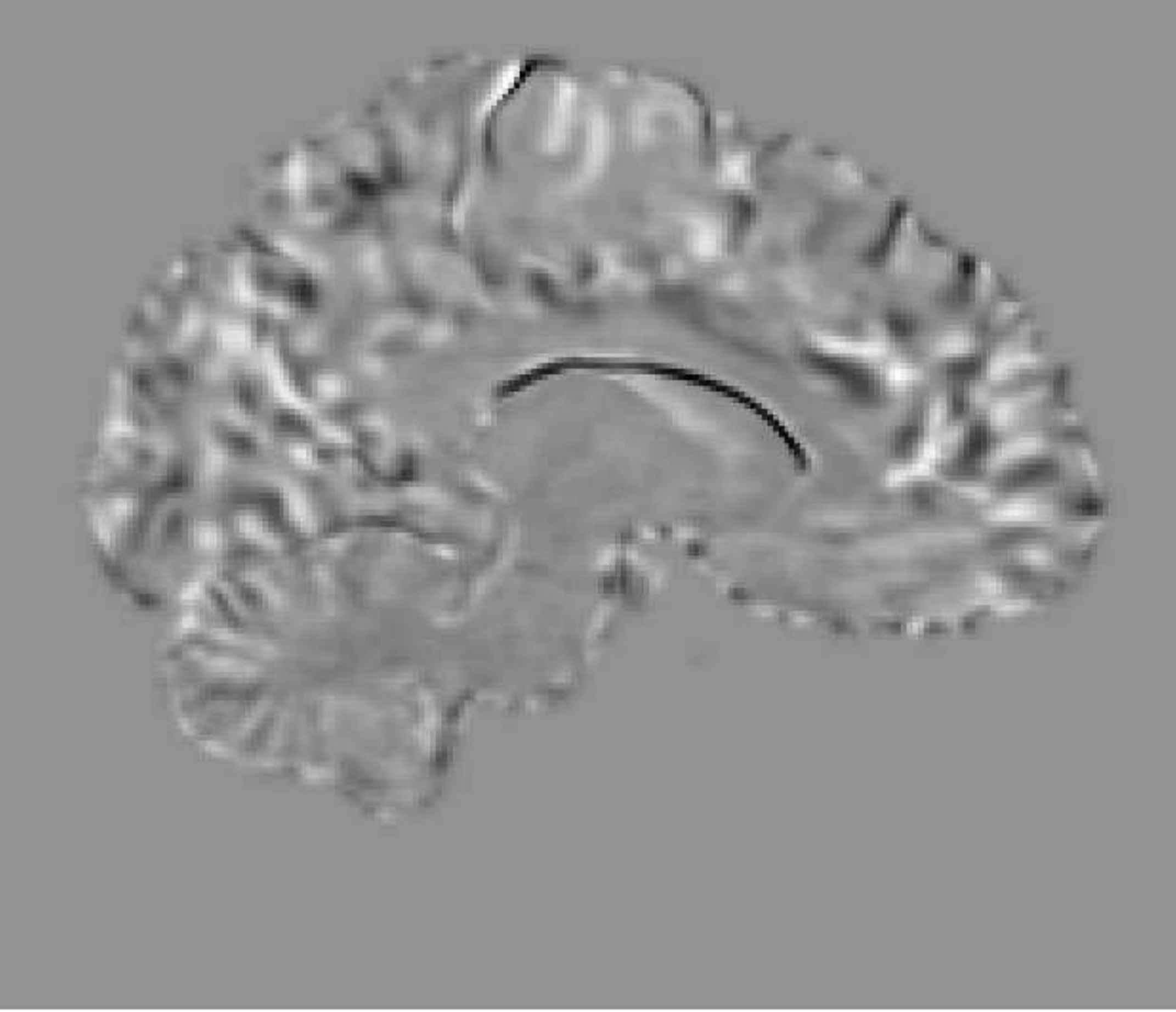} 
&
\includegraphics[angle = 0, width=2.0 cm, height = 1.75 cm]{NStPolzinCNGaussNewton32x32x32_J0Sagittal.pdf} 
&
\includegraphics[angle = 0, width=2.0 cm, height = 1.75 cm]{NStPolzinCNGaussNewton32x32x32_DiffSagittal.pdf} 
\\
\begin{tabular}{c} NSt. BL PDE-LDDMM \\ def. equation, GN \end{tabular} & & \begin{tabular}{c} NSt. BL PDE-LDDMM \\ def. equation, N \end{tabular} & 
\\
\end{tabular}
\caption{ \small Sagittal view of the warped sources and the intensity differences after registration for 
the methods considered in the comparison. $V$-regularization, $\gamma = 0$, BL size of $32 \times 32 \times 32$.} 
\label{fig:3DResults}
\end{figure*}

For a qualitative assessment of the registration methods, we show the registration results obtained by the considered
methods in a selected experiment representative of a difficult deformable registration problem. 
For the methods parameterized in the space of band-limited vector fields, we selected a BL size of 
$32 \times 32 \times 32$.
The methods parameterized in this space show a good compromise between efficiency and accuracy.
Figure~\ref{fig:3DResults} shows the warped images and the difference between the warped and the target image after 
registration.

% --

\subsection{Evaluation}

The evaluation of the methods is based on the accuracy of the registration results for template-based segmentation.
This is a widely extended criterion for non-rigid registration evaluation~\cite{Klein_09,Ou_14}.
We use the manual segmentations provided with the NIREP database as the gold standard.
The Dice Similarity Coefficient (DSC) is selected as the evaluation metric.
Figure~\ref{fig:DSC} shows in the shape of box-and-whisker plots the statistical distribution of the DSC values
obtained after the registration across the 32 segmented structures.
The best performing method was our proposed BL PDE-LDDMM based on the deformation state equation.
The performance of BL PDE-LDDMM based on the state equation was slightly lower and similar to the performance
achieved by St. Mang method.
Gauss-Newton achieved results similar to Newton optimization.
For BL sizes ranging from 64 to 32, all methods performed similarly. 
The performance for BL size of 16 was degraded, indicating that the limit in acceptable BL size selection is above 
this value.

\subsection{Quantitative results for incompressible $H^1$ regularization}

Finally, Table~\ref{table:QuantitativeResultsH1} shows the quantitative results of interest 
for the proposed methods with Gauss-Newton and incompressible $H^1$-regularization.
The model of deformation in inter-subject brain image registration is not compressible.
Therefore, these experiments should be analyzed as a proof of concept regarding the ability
of our methods to compute incompressible velocity fields.
In this case, PDE-LDDMM based on the state equation slightly outperformed PDE-LDDMM on the 
deformation state equation.
All the Jacobian determinants remained equal to 1. 
This shows the ability of our proposed methods to respect the incompressibility constraint.
% Newton optimization converged to $MSE_{rel}$ values above $25 \%$. 

\begin{table}
\scriptsize
\begin{center}
% Benchmark methods, $H^1$-regularization, $\gamma = 1$
% \color{red}
% \begin{tabular}{|c|c|c|c|c|c|c|c|c|}
% \hline
% Method & Opt. & $MSE_{rel}$ & $\Vert g\Vert_{\infty,{rel}}$ & $\max(J(\phi^v))$ & $\min(J(\phi^v))$ & PCG iter & $time_{GPU}(s)$\\
% \hline
% St. Mang & GN & ***19.18 $\pm$ 3.34 & 0.08 $\pm$ 0.05 & 3.71 $\pm$ 0.56 & 0.15 $\pm$ 0.04 & 41.00 $\pm$ 13.02 & 4504.6 $\pm$ \phantom{*}691.5 \\
% NSt. Mang & GN & ***21.11 $\pm$ 5.18 & 0.20 $\pm$ 0.11 & 3.63 $\pm$ 0.83 & 0.14 $\pm$ 0.04 & 25.66 $\pm$ 14.86 & 8688.5 $\pm$ 2045.5\\
% \hline
% \color{black}
% \end{tabular}
\scriptsize
BL PDE-LDDMM, state equation, $H^1$-regularization, $\gamma = 1$
\\
\begin{tabular}{|c|c|c|c|c|c|c|c|}
\hline
BL size & Opt. & $MSE_{rel}$ & $\Vert g\Vert_{\infty,{rel}}$ & $\max(J(\phi^v))$ & $\min(J(\phi^v))$ & PCG iter & $time_{GPU}$ (s) \\
\hline
$64$ & GN & 12.95 $\pm$ 1.36 & 0.07 $\pm$ 0.04 & 1.00 $\pm$ 0.00 & 1.00 $\pm$ 0.00 & 34.00 $\pm$ 1.41 & \phantom{*}422.14 $\pm$ \phantom{**}6.12 \\
\hline
$56$ & GN & 13.59 $\pm$ 1.50 & 0.08 $\pm$ 0.03 & 1.00 $\pm$ 0.00 & 1.00 $\pm$ 0.00 & 32.73 $\pm$ 1.98 & \phantom{*}392.58 $\pm$ \phantom{**}0.82 \\
\hline
$48$ & GN & 14.43 $\pm$ 1.46 & 0.08 $\pm$ 0.03 & 1.00 $\pm$ 0.00 & 1.00 $\pm$ 0.00 & 31.66 $\pm$ 1.71 & \phantom{*}383.49 $\pm$ \phantom{**}4.41 \\
\hline
$40$ & GN & 15.54 $\pm$ 1.70 & 0.07 $\pm$ 0.02 & 1.00 $\pm$ 0.00 & 1.00 $\pm$ 0.00 & 31.13 $\pm$ 2.06 & \phantom{*}388.12 $\pm$ \phantom{*}12.29 \\
\hline
$32$ & GN & 17.44 $\pm$ 1.77 & 0.07 $\pm$ 0.03 & 1.00 $\pm$ 0.00 & 1.00 $\pm$ 0.00 & 30.79 $\pm$ 2.00 & \phantom{*}384.43 $\pm$ \phantom{**}3.20 \\
\hline
$16$ & GN & 27.39 $\pm$ 2.41 & 0.05 $\pm$ 0.02 & 1.00 $\pm$ 0.00 & 1.00 $\pm$ 0.00 & 32.00 $\pm$ 1.06 & \phantom{*}362.62 $\pm$ \phantom{**}4.49 \\
\hline
\end{tabular}
\\
\vspace{0.1 cm}
BL PDE-LDDMM, deformation state equation, $H^1$-regularization, $\gamma = 1$
\\
\begin{tabular}{|c|c|c|c|c|c|c|c|}
\hline
BL size & Opt. & $MSE_{rel}$ & $\Vert g\Vert_{\infty,{rel}}$ & $\max(J(\phi^v))$ & $\min(J(\phi^v))$ & PCG iter & $time_{GPU}$ (s) \\
\hline
$64$ & GN & 13.89 $\pm$ 2.01 & 0.10 $\pm$ 0.04 & 1.00 $\pm$ 0.00 & 1.00 $\pm$ 0.00 & 38.86 $\pm$ 1.68 & \phantom{*}731.81 $\pm$ \phantom{*}15.82 \\
\hline
$56$ & GN & 14.43 $\pm$ 2.21 & 0.11 $\pm$ 5.30 & 1.00 $\pm$ 0.00 & 1.00 $\pm$ 0.00 & 38.00 $\pm$ 2.50 & \phantom{*}535.94 $\pm$ \phantom{*}18.96 \\
\hline
$48$ & GN & 15.16 $\pm$ 2.20 & 0.11 $\pm$ 0.04 & 1.00 $\pm$ 0.00 & 1.00 $\pm$ 0.00 & 37.13 $\pm$ 2.06 & \phantom{*}479.10 $\pm$ \phantom{*}14.55 \\
\hline
$40$ & GN & 16.38 $\pm$ 2.34 & 0.11 $\pm$ 0.05 & 1.00 $\pm$ 0.00 & 1.00 $\pm$ 0.00 & 35.93 $\pm$ 2.12 & \phantom{*}439.13 $\pm$ \phantom{**}2.07 \\
\hline
$32$ & GN & 18.25 $\pm$ 2.70 & 0.11 $\pm$ 0.07 & 1.00 $\pm$ 0.00 & 1.00 $\pm$ 0.00 & 34.73 $\pm$ 3.36 & \phantom{*}407.88 $\pm$ \phantom{**}6.69 \\
\hline
$16$ & GN & 27.89 $\pm$ 2.95 & 0.09 $\pm$ 0.04 & 1.00 $\pm$ 0.00 & 1.00 $\pm$ 0.00 & 37.00 $\pm$ 1.51 & \phantom{*}356.92 $\pm$ \phantom{**}3.96 \\
\hline
\end{tabular}
\end{center}
\caption{ \small Quantitative results of interest for the evaluation of the proposed method.
Results obtained with incompressible $H^1$-regularization and Gauss-Newton-Krylov optimization.
}
\label{table:QuantitativeResultsH1}
\end{table}

\section{Conclusions}
\label{sec:Conclusions}

% - Se ha propuesto 
% - Second-order derivations of Hart et al. and Polzin et al.
% Hessian y Gauss-Newton Hessian approximation
% - Band-Limited formulation

In this work, we have proposed two PDE-constrained LDDMM methods parameterized in the space
of band-limited vector fields and optimized with Newton- and Gauss-Newton- Krylov methods.
The proposed methods depart from two versions of PDE-constrained LDDMM that were optimized using 
gradient information.
We have derived the expressions of the Hessian-vector products and we have computed the
gradient and the Hessian-vector products in the space of band-limited vector fields.
The performance of the proposed methods has been evaluated with respect to the state 
of the art methods most related to our work.

The PDE-constrained LDDMM methods were selected among those that allowed working directly with 
velocity and deformation fields. This circumvented solving the state and adjoint equations 
in the space of band-limited images, which may be subtle for small BL sizes.
It should be noticed that some computations of PDE-LDDMM based on the state equation need to be
performed in the spatial domain, 
while PDE-LDDMM based on the deformation state equation is fully posed in the space of 
band-limited vector fields.

The proposed methods have shown a competitive performance with respect to the benchmark methods.
From them, PDE-LDDMM based on the deformation state equation has shown to outperform the others.
The use of Gauss-Newton optimization and a BL size equal to 32 has shown to be a good configuration 
for obtaining acceptable registration results in an efficient way, while Newton optimization slightly
outperformed these results with a considerable increase of the computational time.
% - H1 experiments, el metodo es capaz de preservar incompressibility constraint.
Indeed, our methods can preserve the incompressibility constraint.

% - Gauss-Newton y Newton comparadas. Similar RMS. 
% Convergencia en state based method no como se espera teoricamente.
% Convergencia en diffeo based method si como se espera teoricamente.
% ...Interesting findings..
% Gauss-Newton has obtained a convergence similar to Newton optimization. 
% The convergence rate in PDE-LDDMM based on the deformation state equation resulted as theoretically expected.
% However, the convergence rate in Gauss-Newton PDE-LDDMM based on the state equation outperformed Newton 
% optimization. ... Citar paper Gauss-Newton, Newton ...
% Computational complexity Gauss-Newton recomendable.

% - Again se reinforce la tesis original de Zhang de BL en LDDMM
The results obtained in this work reinforce Zhang et al.'s idea on the usefulness of working in the space of band-limited
vector fields for LDDMM applications, even in the delicate framework of PDE-constrained LDDMM.
% - Future work, Computational Anatomy applications.
% Explore more physical models, other PDE-LDDMM problems.
In future work, we will analyze the applicability of the proposed methods in Computational Anatomy applications
and we will extend the PDE-LDDMM formulation to more complex physical models.

% - Previous work basado en state equation: para BL sizes pequeñas no daba resultados aceptables
% Este resultado suggests que BL debe ser aplicado en un espacio de vectores, no de imagenes

% \clearpage

\section*{Acknowledgements}
The author would like to give special thanks to Wen Mei Hwu from the University of Illinois 
for interesting ideas in the GPU implementation of the methods.
This work was partially supported by Spanish research grant TIN2016-80347-R.

\bibliographystyle{splncs}
\bibliography{abbsmall.bib,Diffeo.bib,OpticalFlow.bib}
\end{document}